\documentclass[runningheads]{llncs}
\usepackage{graphicx}

\usepackage{tikz}
\usepackage{comment}
\usepackage{amsmath,amssymb} 
\usepackage{color}

\usepackage{tabu}
\usepackage{flushend}
\usepackage{bold-extra}
\usepackage{multirow}

\usepackage{ulem}
\usepackage{nccmath}

\usepackage{floatrow}
\newfloatcommand{capbtabbox}{table}[][\FBwidth]

\usepackage[accsupp]{axessibility}  

\usepackage{booktabs}

\usepackage[pagebackref,breaklinks,colorlinks]{hyperref}
\usepackage[capitalize]{cleveref}
\crefname{section}{Sec.}{Secs.}
\Crefname{section}{Section}{Sections}
\Crefname{table}{Table}{Tables}
\crefname{table}{Tab.}{Tabs.}

\usepackage{multirow}
\usepackage{adjustbox}
\usepackage{bbding}
\usepackage{xspace}
\usepackage[misc]{ifsym}
\usepackage{enumitem}
\setitemize[itemize]{noitemsep,topsep=0in}
\setitemize[enumerate]{noitemsep,topsep=0in}

\definecolor{customdeepgreen}{HTML}{32CD32}
\makeatletter
    \newcommand{\coloruline}[2]{%
        \newcommand\temp@reduline{\bgroup\markoverwith
            {\textcolor{#1}{\rule[-0.5ex]{2pt}{0.8pt}}}\ULon}%
        \temp@reduline{#2}%
    }
    \newcommand{\colorulinea}[2]{%
        \newcommand\temp@redulinea{\bgroup\markoverwith
            {\textcolor{#1}{\rule[-0.5ex]{2pt}{0.8pt}}}\ULon}%
        \temp@redulinea{#2}%
    }
    \newcommand{\colorulineb}[2]{%
        \newcommand\temp@redulineb{\bgroup\markoverwith
            {\textcolor{#1}{\rule[-0.5ex]{2pt}{0.8pt}}}\ULon}%
        \temp@redulineb{#2}%
    }
\makeatother

\makeatletter
\DeclareRobustCommand\onedot{\futurelet\@let@token\@onedot}
\def\@onedot{\ifx\@let@token.\else.\null\fi\xspace}

\def\eg{\emph{e.g}\onedot} 
\def\ie{\emph{i.e}\onedot}

\def\etal{\emph{et al}\onedot}
\makeatother

\newcommand\blfootnote[1]{%
  \begingroup
  \renewcommand\thefootnote{}\footnote{#1}%
  \addtocounter{footnote}{-1}%
  \endgroup
}

\begin{document}
\pagestyle{headings}
\mainmatter
\def\ECCVSubNumber{3606}  

\title{3D Room Layout Estimation from a Cubemap of\\ Panorama Image via\\ Deep Manhattan Hough Transform} 


\titlerunning{3D Room Layout Estimation via DMHT}
%
\author{Yining Zhao\inst{1} \and
Chao Wen\inst{2} \and
Zhou Xue\inst{2} \and
$^\textrm{\Letter}$Yue Gao\inst{1}}

\authorrunning{Y. Zhao et al.}
%
\institute{BNRist, THUIBCS, BLBCI, KLISS, School of Software, Tsinghua University \and Pico IDL, ByteDance}
\maketitle

\begin{abstract}
   Significant geometric structures can be compactly described by global wireframes in the estimation of 3D room layout from a single panoramic image. Based on this observation, we present an alternative approach to estimate the walls in 3D space by modeling long-range geometric patterns in a learnable Hough Transform block. We transform the image feature from a cubemap tile to the Hough space of a Manhattan world and directly map the feature to the geometric output. The convolutional layers not only learn the local gradient-like line features, but also utilize the global information to successfully predict occluded walls with a simple network structure. Unlike most previous work, the predictions are performed individually on each cubemap tile, and then assembled to get the layout estimation. Experimental results show that we achieve comparable results with recent state-of-the-art in prediction accuracy and performance.
   Code is available at \url{https://github.com/Starrah/DMH-Net}.\blfootnote{\noindent$^\textrm{\Letter}$Corresponding author.}


\end{abstract}
\normalem
\section{Introduction}
\label{sec:intro}

\begin{figure*}[!h]
	\centering
	\includegraphics[width=\textwidth]{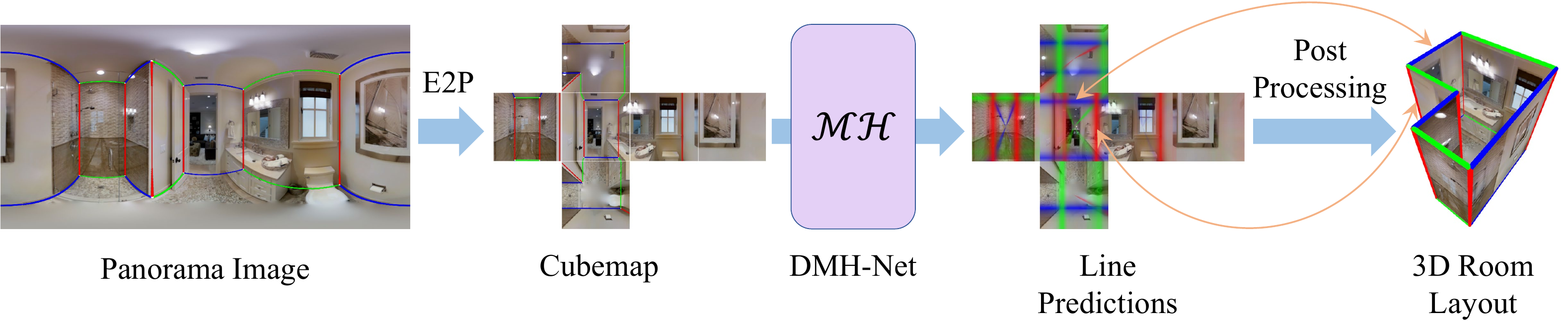}
	\caption{The processing pipeline of our method. Taking panorama image as input, we first apply Equirectangular-to-Perspective transform to generate cubemap, then utilize Deep Manhattan Hough Transform to predict the positions of the wall-wall, wall-floor and wall-ceiling intersection line in each cubemap tile, and recover 3D room layouts by with post-processing procedures.
	}
	\label{fig:pipeline} 
\end{figure*}

Recovering 3D geometry from a single image is one of the most studied stopics in computer vision. This ill-posed task is generally solved under specific scenarios, with certain assumptions or with prior knowledge. The goal of our work is to reconstruct the 3D room layout from a single panorama image under the Manhattan world assumption.

As described in~\cite{bertamini2013visual}, humans perceive 3D world with great efficiency and robustness by using geometrically salient global structures. For our task, an intuitive way of representing the room layout is to use the wireframe for a typical 3D room, consisting of lines on the horizontal plane and vertical lines denoting the junctions between vertical walls. Although using wireframes to estimate 3D room structure is compact and sparse, it could be challenging for vision algorithms to detect long and thin lines with few appearance clues, especially when lines are heavily occluded in a cluttered room.

Existing methods either model the structure estimation as a segmentation problem in panoramic or perspective~\cite{yang2019dula} or decompose the estimation of the geometry elements into a series of regression problems~\cite{sun2019horizonnet}.
The motivation of our work is to take advantage of the compact and efficient representation in the VR area and provide an alternative perspective on this problem. We introduce the cubemap~\cite{ng2005data} of the panorama image and obtain appropriate visual cues in each cubemap tile.
To increase the robustness of the line description, we resort to the Hough transform, which is widely used in line segment detection. It uses two geometric terms, an offset, and an angle, to parameterize lines. As the parameterization of lines is global, the estimation is less prone to be affected by noisy observation and partial occlusions.

In our task, we introduce the Manhattan world assumption to the Hough transform, making the representation even simpler. The wireframe of a room has three types of lines. The first two types are wall-ceiling intersection lines and wall-floor intersection lines, which live on horizontal planes and are perpendicular to each other under the Manhattan world assumption. The third type is vertical lines in the 3D space, which represent the junctions between walls and are perpendicular to the first two types of lines. By taking some pre-processing steps to align the room with the camera coordinate system, the first two types of lines can always be aligned with either the x-axis or the y-axis in the floor/ceiling view.

With the aligned input images, after we adopt the equirectangular to perspective (E2P) transform~\cite{ng2005data,yang2019dula} to get the cubemap from the panorama images, it can be proven that a straight line which is along one of the coordinate axes in the camera 3D space is either a horizontal line, a vertical line or a line passing the center of the image in the cubemap tiles. This simplifies the estimation of the wireframe lines greatly as only lines with specific characteristics are needed for detection so that the network can be more concentrated to learn a better line detection model suitable for Manhattan room layout estimation. 

In this work, we add the Manhattan world line priors into deep neural networks to overcome the challenge of lacking appearance features and occlusion for 3D room wireframes by relying on Hough transform. We embed the Hough transform into trainable neural networks in order to combine Manhattan world wireframe priors with local learned appearances.

The main contributions of this paper can be summarized as follows: 
\begin{itemize}
    \item We introduce the Manhattan world assumption through Deep Hough Transform to capture the long-range geometric pattern of room layouts.
    \item We propose a novel framework estimating layouts on each cubemap tile individually, which is distortion-free for standard CNN.
    \item We directly predict Manhattan lines with explicit geometric meaning, which achieves comparable performance towards recent state-of-the-art works.
\end{itemize}

\section{Related Work}
\noindent{\textbf{Room layout estimation.}}
3D room layout estimation from a single image attracted a lot of research over the past decade. Most previous studies exploit the Manhattan world assumption~\cite{coughlan1999manhattan} which means that all boundaries are aligned with the global coordinate system. Moreover, vanishing points detection can be used to inferring the layout based on the assumption.

Traditional methods extract geometric cues and formalize this task as an optimization problem. Since the images may differ in the FoV (ﬁeld of view), ranging from perspective to $360^\circ$ panoramas, the methods vary with types of input images.
In terms of perspective images, Delage \etal~\cite{delage2006dynamic} propose a dynamic Bayesian network model to recognize ``floor-wall'' geometry of the indoor scene. Lee \etal~\cite{lee2009geometric} using Orientation Map (OM) while Hedau \etal~\cite{hedau2009recovering} using Geometric Context (GC) for geometry reasoning to tackle the problem. The strategies have been employed by other approaches leveraging enhanced scoring function~\cite{schwing2012efficient,schwing2012efficient2}, or modeling the objects-layout interaction~\cite{del2013understanding,gupta2010estimating,zhao2013scene}.

On the other hand, since the $360^\circ$ panoramas provides more information, there are multiple papers exploiting in this direction. Zhang \etal~\cite{zhang2014panocontext} proposes to estimate layout and 3D objects by combining the OM and GC on a panoramic image.
Yang \etal~\cite{yang2016efficient} takes line segments and superpixel facets as features and iteratively optimize the 3D layout.
Xu \etal~\cite{xu2017pano2cad} estimate layout use detected objects, their pose, and context in the scene.
In order to recover the spatial layout, Yang \etal~\cite{yang2018automatic} use more geometric cues and semantic cues as input, while Pintore \etal~\cite{pintore2016omnidirectional} utilize the gradient map.

With the astonishing capability, neural network based methods leverage data prior to improve layout estimation. Most studies in this field focused on dense prediction, which train deep classification network to pixel-wise estimate boundary probability map~\cite{mallya2015learning,ren2016coarse,zhao2017physics}, layout surface class~\cite{dasgupta2016delay,izadinia2017im2cad} or corner keypoints heatmaps~\cite{lee2017roomnet,fernandez2020corners}. 
The panoramas-based approach has recently attracted wide interest to 3D room layout estimation problems. Zou \etal~\cite{zou2018layoutnet} predict the corner and boundary probability map directly from a single panorama. Yang \etal~\cite{yang2019dula} propose DuLa-Net which leverages both the equirectangular view and the perspective floor/ceiling view images to produce 2D floor plan. Pintore \etal~\cite{pintore2020atlantanet} follows the similar approach, which directly adopts equirectangular to perspective (E2P) conversion to infer 2D footprint. Fernandez-Labrador \etal~\cite{fernandez2020corners} present EquiConvs, a deformable convolution kernel which is specialized for equirectangular image. Zeng \etal~\cite{zeng2020joint} jointly learns layout prediction and depth estimation from a single indoor panorama image. Wang \etal~\cite{Wang_2021_LED2Net} propose a differentiable depth rendering procedure which can learn depth estimation without depth ground truth.
Although extensive research has been carried out on 3D room layout estimation, few studies exist which try to produce a more compact representation to infer the layout. Sun \etal propose HorizonNet~\cite{sun2019horizonnet} and HoHoNet~\cite{Sun_2021_HoHoNet}, which encodes the room layout as boundary and corner probability vectors and propose to recover the 3D room layouts from 1D predictions. 
LGT-Net~\cite{jiang2022lgt} is a recent work which represents the room layout by horizon-depth and room height. NonCuboidRoom~\cite{yang2022learning} takes lines into account and recovers partial structures from a perspective image, but cannot estimate the whole room.
Although these existing methods are sound in predicting layout, they still have limitations in representing layout.
Our approach also assembles compact results but is different from the existing methods. Rather than regressing boundaries values, our approach combines the cubemap and parametric representation of lines.

\noindent{\textbf{Hough transform based detectors.}} Hough~\etal~\cite{hough1962method} devise Hough transform to detect line or generalized shapes~\cite{ballard1981generalizing} \eg circle from images. Through the extensive use of Hough transform, traditional line detectors apply edge detection filter at the beginning (\eg Canny~\cite{canny1986computational} and Sobel~\cite{sobel19683x3}), then identify signiﬁcant peaks in the parametric spaces. Qi~\etal~\cite{qi2019deep} using the Hough voting schemes for 3D object detection. Beltrametti~\etal~\cite{beltrametti2020geometry} adopt voting conception for curve detection.
Recently, learning-based approaches demonstrate the representation capability of the Hough transform. Han~\etal~\cite{han2020deep} propose Hough Transform to aggregate line-wise features and detect lines in the parametric space. Lin~\etal~\cite{lin2020deep} consider adding geometric line priors through Hough Transform in networks. Our work use Deep Hough features in a new task to tackle the room layout estimation problem.
\section{Method}
\subsection{Overview}

Our goal is to estimate the Manhattan room layout from a single $360^\circ$ panoramic image. However, panoramic images have distortion, \ie a straight line in the 3D space may not be straight in the equirectangular view of panoramic images. Instead of make predictions directly on the panoramic image like previous works \cite{zou2018layoutnet,sun2019horizonnet,Wang_2021_LED2Net,Sun_2021_HoHoNet}, we get a cubemap~\cite{ng2005data} which contains six tiles by adopting E2P transform~\cite{ng2005data,yang2019dula}.

Given a single RGB panoramic image as input, first we take some pre-processing steps to align the image, get the cubemap and transform ground truth labels.
Then we use our proposed Deep Manhattan Hough Network (DMH-Net) to detect three types of straight lines on the tiles of the cubemap. 
Finally, with an optimization-based post-processing procedure, the line detection results can be assembled and fully optimized to generate the 3D room layout.

In \cref{sec:pre}, we introduce our pre-processing procedure.
The Deep Manhattan Hough Transform for room layout estimation is presented in \cref{sec:manhattanhough}.
We summarize the network architecture of the proposed DMH-Net in \cref{sec:arch}.
Finally, in \cref{sec:post}, we introduce our optimization-based post-processing method.

\subsection{Pre-processing}\label{sec:pre}
\noindent{\textbf{Aligning the Image.}} 
Receiving a single panorama which covers a $360^\circ$ H-FoV, we first align the image based on the LSD algorithm and vanishing points mentioned in~\cite{zou2018layoutnet,sun2019horizonnet}. Our approach exploits both the Manhattan world assumption and the properties of aligned panoramas. After the alignment, the cubemap tiles are aligned with the three principal axes~\cite{coughlan1999manhattan}, \ie, the optical axis of the front camera is perpendicular to a wall. Aligning the panorama makes the vanishing point precisely the center of each cubemap tile.


\noindent{\textbf{Cubemap Generation.}} 
E2P transform is conducted six times on the equalrectangular image with different azimuth angle and elevation angle to generate cubemap tiles $I_{front}, I_{back}, I_{left}, I_{right}, I_{ceil}, I_{floor}$, as Fig.~\ref{fig:pipeline} shown. For 
all of the cubemap tiles, FoV is set to $90^\circ$ on both horizontal and vertical direction, and image size is set to $512\times 512$.

\noindent{\textbf{Ground Truth Transformation.}}
After adopting the aligning method described above for all the datasets we use, the ground truth of the room’s layout is provided in the format of corner coordinates in the panoramic image.
Using the same E2P transform as described previously, we can also transform the coordinates of the ground truth corner coordinate from the panoramic image to the cubemap and then get lines in the cubemap by connecting the points by their connectivity relation in the original panoramic image.
The lines can be categorized into three types: horizontal lines, vertical lines and lines passing the center of the image. 
Since Deep Manhattan Hough Transform can only detect straight lines rather than line segments, we do not care about the specific position of the line's start point and end point. So for a horizontal (or vertical) line, we only use the $y$ (or $x$) coordinate to represent the line, and for a line passing the center, we only use the orientation angle $\theta$ to represent it.



\subsection{Deep Manhattan Hough Transform}\label{sec:manhattanhough}
In the case of line detection, the traditional Hough Transform~\cite{hough1962method} parameterizes lines in images to polar coordinates with two parameters, an orientation $\theta$ and a distance $\rho$. Every image pixel votes to a discretized parameter bin, which represents the corresponding polar coordinates. The parameter space is denoted as the Hough space, and the maximum local peaks represent the lines in the image. 
Specifically, given an single channel input $\mathbf{X} \in \mathbb{R}^{h_0\times w_0}$, the Hough Transform $\mathcal{H}$ can be represented as:
\begin{equation}\label{formula:traditionalHT}
    \mathcal{H}(\rho, \theta) = \sum_{i\in l} \mathbf{X}(x_{i}, y_{i})
\end{equation}
in which $l$ is a line whose orientation angle is $\theta$ and distance to the coordinate origin is $\rho$, and $i$ is each of the point in line $l$. 

A key concept of our network is detecting all possible positions of the layout boundaries of the room in each cubemap tile.
We propose to incorporate the deep network with Hough Transform for layout boundaries detection. Specifically, we present the Deep Manhattan Hough Transform (DMHT) combining the deep CNN feature and Manhattan world assumption. It is based on the following two assumption: 
\begin{enumerate}
    \item The Manhattan world assumption, \ie, all of the walls, the ceiling and the floor must be perpendicular to each other, and all of the intersection lines of them must be parallel with one of the coordinate axes of some orthogonal coordinate space (named Manhattan space).
    \item The input image must be aligned, \ie, the camera of each cubemap tile faces precisely to one of the walls, and its optical axis is parallel with one of the coordinate axes of the Manhattan space. 
\end{enumerate}

In practice, the two assumptions are quite straightforward since most of the rooms in human buildings obey the Manhattan world assumption, and the second assumption can be implemented with the pre-processing steps described in \cref{sec:pre}. Under these two assumptions, it can be proven that any line in the wireframe of the room, including wall-wall line, wall-ceiling line and wall-floor line, must be either a horizontal line ($\theta=0$), a vertical line ($\theta=\pi/2$) or a line passing the center ($\rho=0$) in the cubemap tiles, which is a special case of single-point perspective~\cite{horry1997tour}~(more details in supplementary materials).
 
\begin{figure}[!h]
	\centering
	\includegraphics[width=0.5\columnwidth]{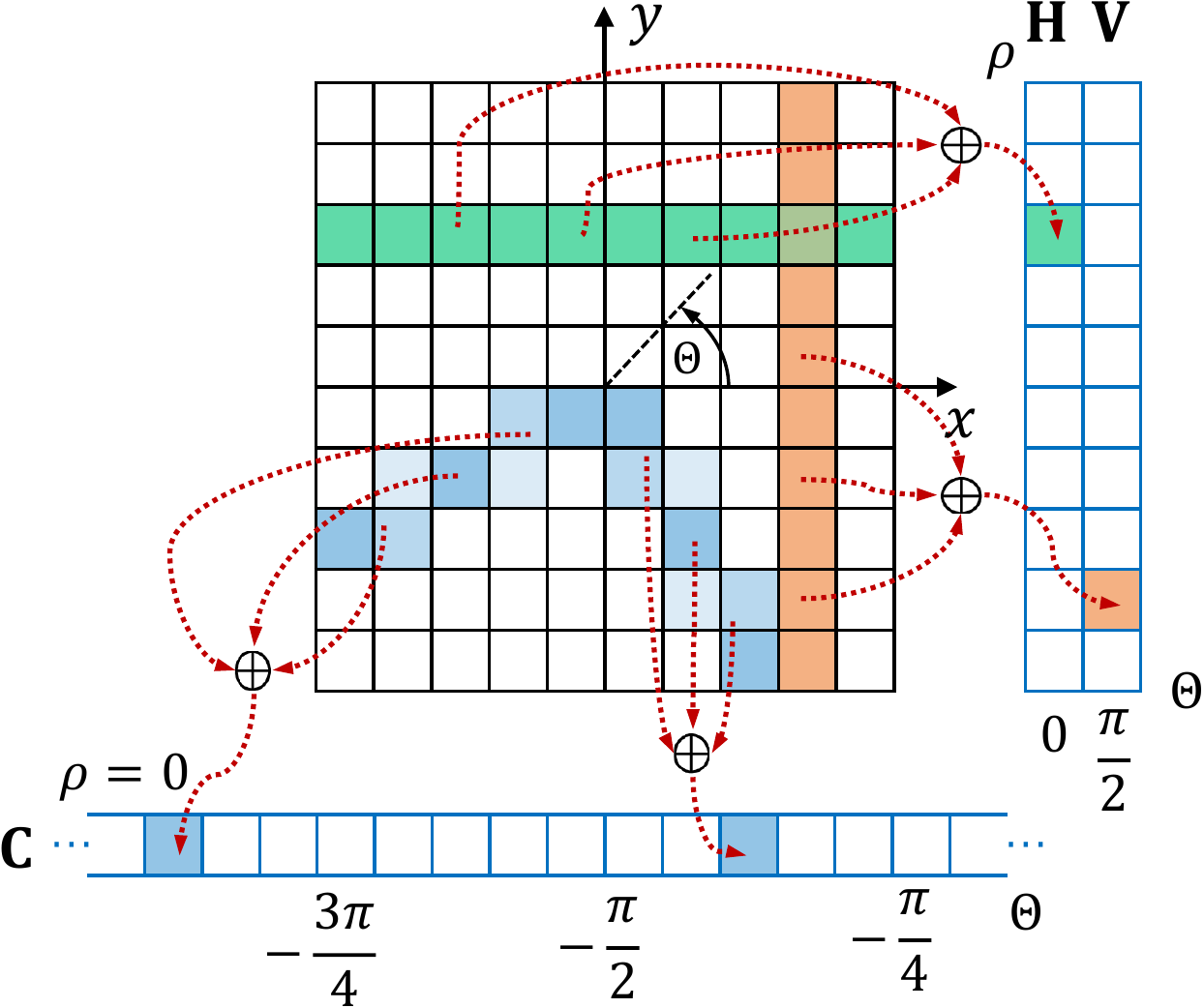}
	\caption{\textbf{Deep Manhattan Hough Transform Overview. } Each feature map of the CNN can be regarded as many discretized bins. According to the Hough Transform, the bins with the same polar coordinate parameters vote for the same single bin in the Hough space. A bin in the vector $\mathbf{H}$ (green) is calculated by aggregating horizontal features in the 2D feature map, while bins in $\mathbf{V}$(orange) and $\mathbf{C}$(blue) are calculated by aggregate vertical features and features passing the center respectively.}
	\label{fig:hough} 
\end{figure}

As shown in Fig.~\ref{fig:hough}, given the feature maps of a cubemap tile extracted by the encoder network as input, for each channel $\mathbf{X} \in \mathbb{R}^{h\times w}$ of the feature maps, the Deep Manhattan Hough Transform $\mathcal{MH}$ output three vectors $\mathbf{H}, \mathbf{V}$ and $\mathbf{C}$, corresponding to bins in the Hough space representing horizontal lines, vertical lines and lines passing the center, which is defined as:
\begin{equation}\label{formula:DMHT}\begin{aligned}
    \mathbf{H}(\rho)&=\mathcal{MH}_{H}(\rho) = \mathcal{H}(\rho, 0) = \sum_{x_i=-\frac{w}{2}}^{\frac{w}{2}} \mathbf{X}(x_i, \rho),\\
    \mathbf{V}(\rho)&=\mathcal{MH}_{V}(\rho) = \mathcal{H}(\rho, \frac{\pi}{2}) = \sum_{y_i=-\frac{h}{2}}^{\frac{h}{2}} \mathbf{X}(\rho, y_i),\\
    \mathbf{C}(\theta)&=\mathcal{MH}_{C}(\theta) = \mathcal{H}(0, \theta)\\
    &=\left\{ \begin{array}{lr}
        \sum_{x_i=0}^{\frac{w}{2}} \mathbf{X}(x_i, x_i\cdot \tan(\theta)), &|\tan(\theta)| \leq 1 \\
        \sum_{y_i=0}^{\frac{h}{2}} \mathbf{X}(y_i \cdot \cot(\theta), y_i), &|\tan(\theta)| > 1
    \end{array}\right.
\end{aligned}\end{equation}
in which $\rho \in [-\frac{h}{2},\frac{h}{2}]$ for $\mathbf{H}$ and $\rho \in [-\frac{w}{2},\frac{w}{2}]$ for $\mathbf{V}$, and $0\leq\theta\leq 2\pi$. $\mathbf{H}(\rho)$ is the bin of $\mathbf{H}$ with Hough space parameter $\rho$, and similar for $\mathbf{V}$ and $\mathbf{C}$. 

To calculate the proposed DMHT efficiently, effective discretization is necessary. It is natural to discretize $\rho$ to be integers so that each bin in $\mathbf{H}$ and $\mathbf{V}$ represents to a 1-pixel-wide line in the image, thus $\mathbf{H}\in \mathbb{R}^{h}$ and $\mathbf{V}\in \mathbb{R}^{w}$. In our experiment, we discretize $\theta$ so that $\mathbf{C}\in \mathbb{R}^{2(h+w)}$for each bin in $\mathbf{C}$, the corresponding line intersects the image's border at positions whose coordinates are integer. With the discretization technique above, the process of DMHT can be implemented with matrix addition and multiplications, which is highly parallelizable and suitable for GPU calculation. See the supplementary materials for more detail.

\begin{figure*}[!h]
	\centering
	\includegraphics[width=\textwidth]{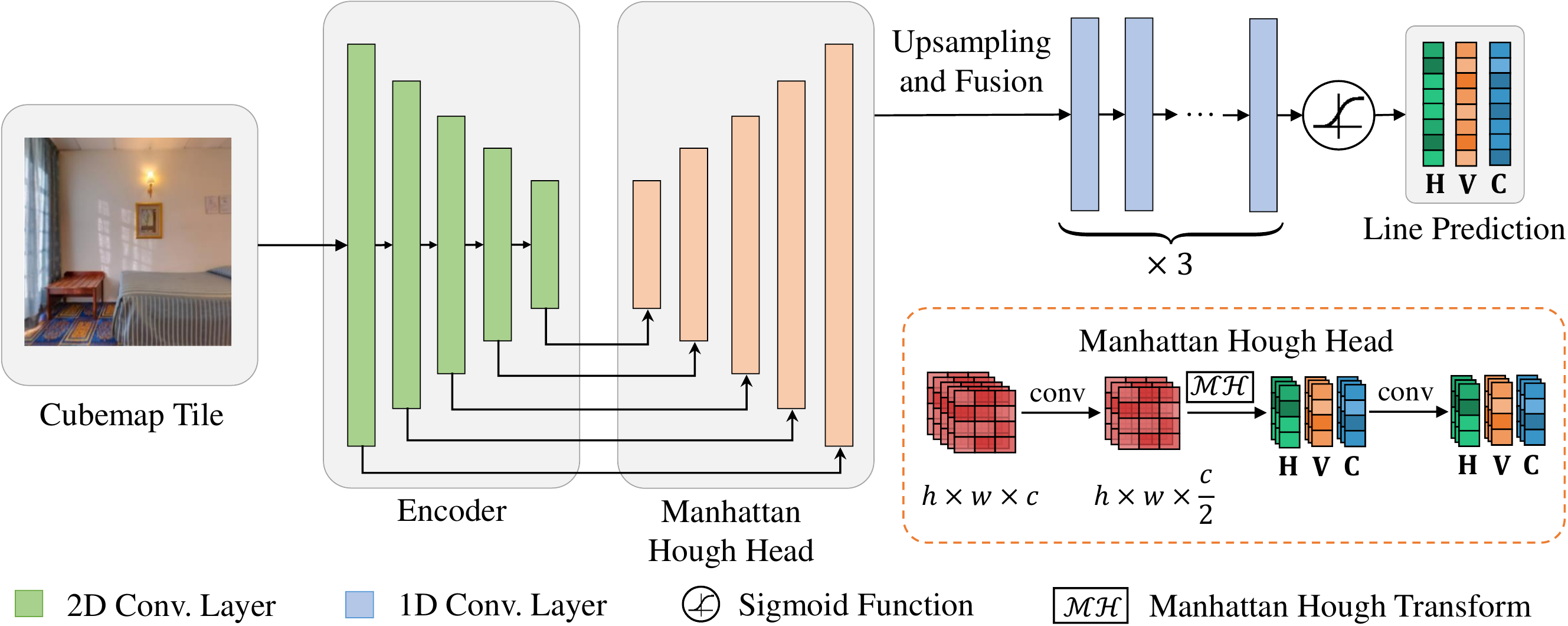}
	\caption{Overview of the architecture of Deep Manhattan Hough Network(DMH-Net). Given a cubemap tile image as input, a CNN encoder is adopted to extract multi-level image features. Manhattan Hough Head are utilized to handle multi-scale 2D features and perform the Manhattan Hough Transform to get three feature vectors in the Hough space, $\mathbf{H}\in\mathbb{R}^{h\times \frac{c}{2}}$, $\mathbf{V}\in\mathbb{R}^{w\times \frac{c}{2}}$ and $\mathbf{C}\in\mathbb{R}^{2(h+w)\times \frac{c}{2}}$. Finally, the feature vectors are fused to generate lines prediction result, in the format of line position probability vector.
	}
	\label{fig:model} 
\end{figure*}

\subsection{Network Architecture}\label{sec:arch}
\noindent{\textbf{Feature Extractor.}}
We employ Dilated Residual Networks~\cite{yu2017dilated,He_2016_ResNet} as our encoder, which utilizes dilated convolution to imporve spatial acuity for the images to learn better feature for thin line prediction. The input shape of panorama in equirectangular view $I_{equi}$ is $512 \times 1024 \times 3$.
For each of the six tiles of the cubemap, the input shape is $512 \times 512 \times 3$.
To capture both low-level and high-level features, we extract the intermediate features of the encoder network. In particular, we gather the perceptual feature from each block before downsample layers. Then, all the features are fed into five independent \textbf{Manhattan Hough Head} with different input and output size. 

\noindent{\textbf{Manhattan Hough Head.}}
As shown in Fig.~\ref{fig:model}, Manhattan Hough Head receives feature maps of cubemap tile and produces feature vectors in Hough space for line detection. Specifically, given input feature maps with channel number $c$, the module first applies 2D convolution to reduce the feature channels to $\frac{c}{2}$, then transform the feature through DMHT described in \ref{sec:manhattanhough} to get results in Hough space. The results are then filtered by a 1D convolution layer with kernel size $=3$ to capture context information surrounding certain Hough bins. The output of this module is three multi-channel feature vectors in the Hough space: $\mathbf{H}\in \mathbb{R}^{h\times \frac{c}{2}}$, $\mathbf{V}\in \mathbb{R}^{w\times \frac{c}{2}}$ and $\mathbf{C}\in \mathbb{R}^{2(h+w)\times \frac{c}{2}}$. 

\noindent{\textbf{Upsampling, Fusion and Generating Line Prediction.}}
Since the size of the feature maps extracted by the feature extractor varies with the depth of the layer, the size of the output feature vectors of the five Manhattan Hough Heads differs. We upsample all the feature vectors to the same size which equals to the width and height of the original image, $h=512$ and $w=512$, with bilinear interpolation. Then, feature vectors of the same type are fused by channel-wise concatenation, and filtered by three 1D convolution layers.
Finally, the Sigmoid function is applied to generate three single-channel 1D vectors $\mathbf{H}\in\mathbb{R}^{h}$, $\mathbf{V}\in\mathbb{R}^{w}$ and $\mathbf{C}\in\mathbb{R}^{2(h+w)}$, which represents the predicted probability of horizontal lines, vertical lines and lines passing the center of the image. 

\noindent{\textbf{Loss Functions.}}
If we define the ground truth of the probability vectors $\mathbf{H}, \mathbf{V}$ and $\mathbf{C}$ as binary-valued vectors with 0/1 labels simply, it would be too sparse to train, \eg there would be only less than two non-zero values out of 512 in a cubemap tile.
Similar as HorizonNet\cite{sun2019horizonnet}, we smooth the ground truth based on the exponential function of the distance to nearest ground truth line position. 

Then, for $\mathbf{H}$, $\mathbf{V}$ and $\mathbf{C}$, we can apply binary cross entropy loss:
\begin{equation}
\mathcal{L}_{bce}(\mathbf{X}, \mathbf{X}^*) = - \sum_{i} x_i^* \log(x_i) + (1 - x_i^*)\log(1-x_i)\\
\end{equation}
in which $x_i$ denotes the $i$-th element of $\mathbf{X}$.

The overall loss for layout estimation is defined by the sum of the loss of the three type of lines:
\begin{equation}\begin{aligned}
\mathcal{L} = \mathcal{L}_{bce}(\mathbf{H}, \mathbf{H}^*) + \mathcal{L}_{bce}(\mathbf{V}, \mathbf{V}^*) + \mathcal{L}_{bce}(\mathbf{C}, \mathbf{C}^*)
\end{aligned}\end{equation}
where the notation $*$ indicates ground truth data.

\subsection{Post-processing}\label{sec:post}
Since the output of our network is line existence confidence on the cubemap, post-processing is necessary to generate the final 3D layout result. Our post-processing procedure consists of two stages: initialization and optimization~(more details in supplementary materials).

\noindent{\textbf{Parametric Room Layout Representation.}}
In 3D space, take the camera as the coordinate origin and z-axis points to the ceiling, and regard the distance from the camera to the floor can be set as a known value similar to Zou~\etal~\cite{zou2018layoutnet}. The layout of a Manhattan room with $n$ vertical walls can be represents with $n+1$ parameters: $n$ for the distance on coordinate axis from the camera to each wall, the other for the height of the room. 

\noindent{\textbf{Layout Parameters Initialization.}}
Given the lines predictions in the cubemap, we can calculate the elevation angle of each wall-ceiling and wall-floor intersection line, and since the distance from the camera to the floor is known, the distance to each of the wall-floor lines can be directly estimated, which is used as initial values for the first $n$ parameters of the layout representation. Though the distance from the camera to the ceiling is unknown, by assuming and optimizing the ratio of the distance from the camera to the ceiling and the floor so that the ceiling 2D frame has the highest 2DIoU with the floor 2D frame, we can get the initial estimated value of the room height. 

\noindent{\textbf{Gradient Descent Based Optimization.}}
Given the initialized layout parameters, we convert the them into corner point coordinates in the panoramic image and then line positions in the cubemap. For each line position in the cubemap tiles, we can get line existence confidence from the output $H$, $V$ and $C$ of the network. We add the confidence of all the line position together to get a score representing the overall confidence of the layout parameters, define the loss as the negative of the score and optimize the loss using SGD, with learning rate 0.01, no momentum and 100 optimization steps. With the optimization process, the line predictions on different tiles of the cubemap can be effectively integrated together to get a better layout estimation from the global aspect.
\section{Experiments}
\subsection{Experimental setup}
\noindent\textbf{Dataset.} We use three different datasets. The PanoContext dataset~\cite{zhang2014panocontext} and the Stanford 2D-3D dataset~\cite{armeni2017joint,zou2018layoutnet} are used for cuboid layout estimation, while the Matterport 3D dataset~\cite{Matterport3D} is used for non-cuboid Manhattan layout estimation. For fair comparison, we follow the train/validation/test split adopted by other works~\cite{zou2018layoutnet,sun2019horizonnet,zou2021_layoutv2}.

\noindent\textbf{Evaluation Metric}
We employ standard evaluation metrics for 3D room layout estimation. For both cuboid and non-cuboid estimation, we use \textbf{(i) 3D IoU}, which is defined as the volumetric intersection over the union between the prediction and the ground truth. In addition, for cuboid layout estimation, we also use \textbf{(ii) Corner Error} (CE), which measures the average Euclidean distance between predicted corners and ground truth, and \textbf{(iii) Pixel Error} (PE), which is the pixel-wise accuracy. For non-cuboid estimation, we also use \textbf{(iv) 2D IoU}, which project 3D IoU to the 2D plane, and \textbf{(v) $\delta_{i}$}, which is defined as the percentage of pixels where the ratio between the prediction and the ground truth is no more than 1.25. For 3D IoU, 2DIoU and $\delta_{i}$, the larger is better. For Corner Error and Pixel Error, the smaller is better. 

\noindent\textbf{Baselines}
We compare to previous works for both cuboid and non-cuboid room layout estimation and evaluate quantitative performance.
The methods we compare to include PanoContext~\cite{zhang2014panocontext}, LayoutNet~\cite{zou2018layoutnet}, CFL~\cite{fernandez2020corners}, DuLa-Net~\cite{yang2019dula}, HorizonNet~\cite{sun2019horizonnet}, AtlantaNet~\cite{pintore2020atlantanet}, HoHoNet\cite{Sun_2021_HoHoNet} and LED$^2$-Net~\cite{Wang_2021_LED2Net}. 
In addition, we also compared the improved version of the algorithm proposed in the recent work of Zou~\etal~\cite{zou2021_layoutv2}, \ie, Layoutnet-v2, DuLa-Net-v2 and HorizonNet$^+$.
We compare with all these methods based on published comparable results.
More specifically, DuLa-Net~\cite{yang2019dula} and CFL~\cite{fernandez2020corners} take $256\times512$ resolution images as input while others using the input resolution of $512\times1024$.
Additionally, PanoContext~\cite{zhang2014panocontext} and LayoutNet~\cite{zou2018layoutnet} need additional input (\ie Line Segment or Orientation Map) besides a single RGB panorama image.

\noindent\textbf{Implementation Details}\label{sec:implemention_details}
The network is implemented in PyTorch and optimized using Adam~\cite{kingma2014adam}. The network is trained for 75 epochs with batch size 8 and learning rate 1e-4.
The model is trained on a single NVIDIA A100 GPU. 
For both cuboid and non-cuboid estimation, we use the Pano Stretch\cite{sun2019horizonnet}, left-right flipping for data augmentation. In addition, for non-cuboid estimation, we also apply vertical flip and random $90^{\circ}$ rotation augmentation on the cubemap tiles generated by the E2P transform.

\subsection{Cuboid Room Results}\label{sec:cuboid_result}

\begin{table}[t]
\centering
\resizebox{.7\textwidth}{!}{%
\begin{tabu} to 0.85\textwidth {X[3.5c]X[c]X[c]X[c]X[c]X[c]X[c]}
\toprule
\multirow{2}{*}{Method}  & \multicolumn{3}{c}{PanoContext}  & \multicolumn{3}{c}{Stanford 2D-3D} \\
\cmidrule(lr){2-4} \cmidrule(lr){5-7} 
& 3DIoU  & CE & PE & 3DIoU  & CE & PE  \\
\midrule
PanoContext~\cite{zhang2014panocontext} & 67.23 &  1.60 & 4.55 & - & - & -\\
LayoutNet~\cite{zou2018layoutnet}  & 74.48    & 1.06    & 3.34 & 76.33 & 1.04 & 2.70\\ 
DuLa-Net~\cite{yang2019dula}     & 77.42    & - & -  & 79.36 & - & -\\ 
CFL~\cite{fernandez2020corners}    & 78.79    & 0.79    & 2.49 & - & - & -\\ 
HorizonNet~\cite{sun2019horizonnet}   & 82.17    & 0.76    & 2.20 & 79.79 & 0.71 & 2.39 \\ 
AtlantaNet~\cite{fernandez2020corners} & - & - & - & 82.43 & 0.70 & 2.25 \\ 
LED$^2$-Net~\cite{Wang_2021_LED2Net}   & 82.75    & -    & -  & 83.77 & - & - \\ 
\textbf{DMH-Net (Ours)} & \textbf{85.48} & \textbf{0.73} & \textbf{1.96} & \textbf{84.93} & \textbf{0.67} & \textbf{1.93}\\
\bottomrule
\end{tabu}
}
\caption{Quantitative results of cuboid room layout estimation evaluated on the testset of the PanoContext dataset~\cite{zhang2014panocontext} and the Stanford 2D-3D dataset~\cite{armeni2017joint,zou2018layoutnet}. CE means Corner Error, and PE means Pixel Error.}
\label{tab:panocontext}
\end{table}

\begin{figure}[!b]
	\centering
	\includegraphics[width=0.86\textwidth]{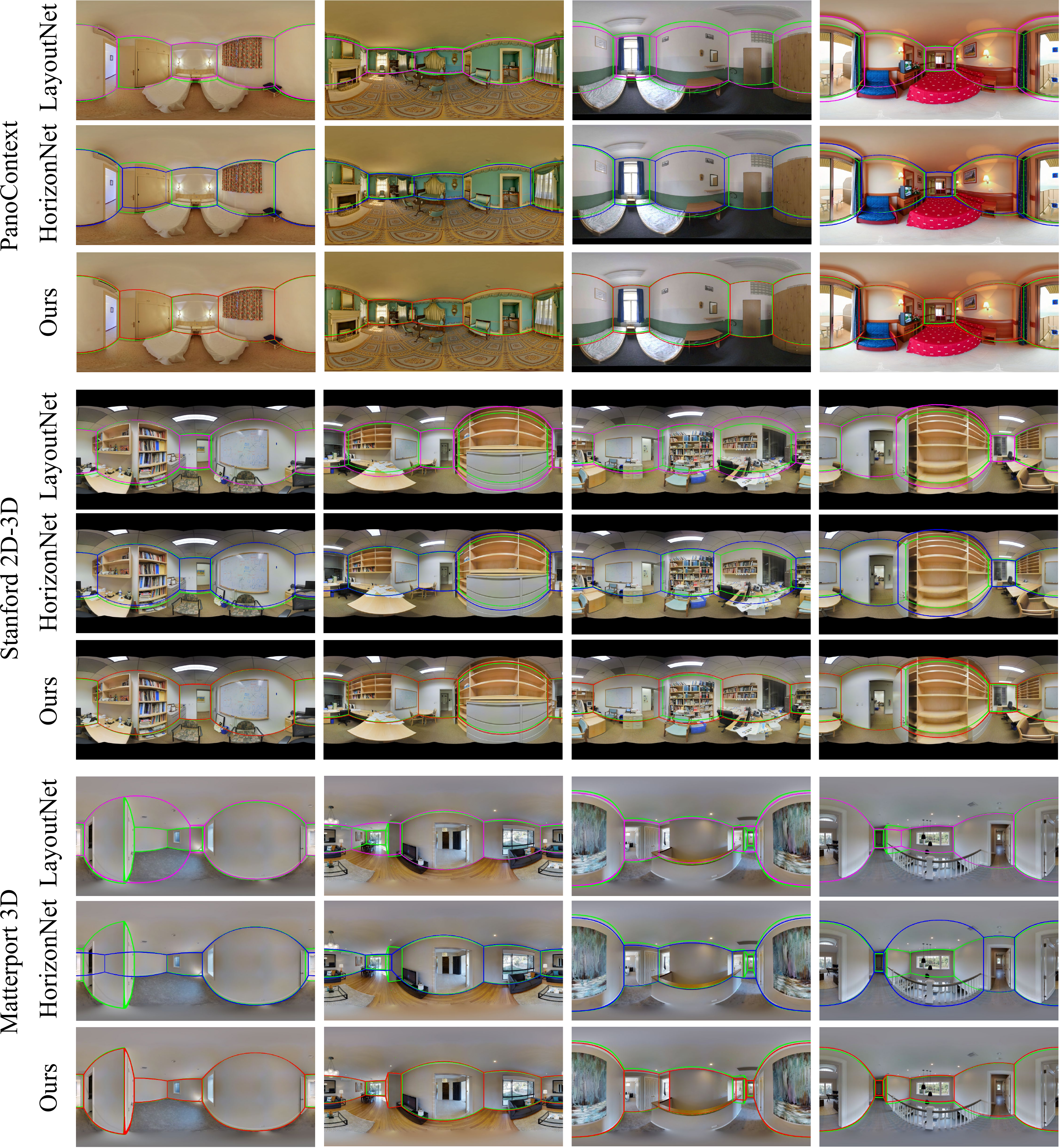}
	\caption{Qualitative results of both cuboid and non-cuboid layout estimation. The results are selected from four folds that comprise results with the best 0–25\%, 25–50\%, 50–75\% and 75–100\% 3DIoU (displayed from left to right). The green lines are ground truth layout while the pink, blue and orange lines are estimated by LayoutNet~\cite{zou2018layoutnet}, HorizonNet~\cite{sun2019horizonnet} and our DMH-Net.
    }
\label{fig:visual_result} 
\end{figure}

Tab.~\ref{tab:panocontext} show the quantitative comparison on the PanoContext dataset~\cite{zhang2014panocontext} and Stanford 2D-3D dataset~\cite{armeni2017joint,zou2018layoutnet}, respectively. Blank fields in the tables indicate metrics which is not reported in the corresponding publications.
As can be seen, our approach achieves the state-of-the-art results under these two datasets in cuboid 3D room layout estimation. 
The qualitative results of the PanoContext and the Stanford 2D-3D dataset are shown in \cref{fig:visual_result}.

\subsection{Non-cuboid Room Results}\label{sec:non-cuboid_result}
\cref{tab:matter} shows the quantitative comparison on the Matterport 3D dataset~\cite{Matterport3D}. Blank fields in the table indicate metrics which is not reported in the corresponding publications.
As can be seen, our approach achieves competitive overall performance with the state-of-the-art methods in non-cuboid 3D layout estimation.
The qualitative results of the Matterport 3D dataset are shown in \cref{fig:visual_result}. We present results from different methods aligned with ground truth. Please see supplemental materials for more qualitative results.

\begin{table*}[]
\centering
\resizebox{\textwidth}{!}{%
\begin{tabular}  {p{4cm}ccccccccccccccc}
\toprule
\multicolumn{1}{c}{Metrics}       & \multicolumn{5}{c}{3DIoU}         & \multicolumn{5}{c}{2DIoU}         & \multicolumn{5}{c}{$\delta_{i}$}      \\ 
\cmidrule(lr){1-1} \cmidrule(lr){2-6} \cmidrule(lr){7-11} \cmidrule(lr){12-16} 
\multicolumn{1}{c}{\# of corners} & 4   & 6   & 8   & 10+ & Overall        & 4   & 6   & 8   & 10+ & Overall        & 4   & 6   & 8   & 10+ & Overall        \\ 
\midrule
\multicolumn{1}{c}{LayoutNet-v2~\cite{zou2021_layoutv2}}  & 81.35          & 72.33          & 67.45          & 63  & 75.82          & 84.61          & 75.02          & 69.79          & 65.14          & 78.73          & 0.897          & 0.827          & 0.877          & 0.8 & 0.871          \\ 
\multicolumn{1}{c}{DuLa-Net-v2\cite{zou2021_layoutv2,yang2019dula}}   & 77.02          & 78.79          & 71.03          & 63.27          & 75.05          & 81.12          & 82.69          & 74  & 66.12          & 78.82          & 0.818          & 0.859          & 0.823          & 0.741          & 0.818          \\ 
\multicolumn{1}{c}{HorizonNet$^+$\cite{zou2021_layoutv2,sun2019horizonnet}}   & 81.88          & 82.26          & 71.78          & 68.32          & 79.11          & 84.67          & 84.82          & 73.91          & 70.58          & 81.71          & 0.945          & 0.938          & 0.903          & 0.861          & 0.929          \\ 
\multicolumn{1}{c}{AtlantaNet\cite{pintore2020atlantanet}}    & 82.64          & 80.1& 71.79          & \textbf{73.89} & \textbf{81.59} & 85.12          & 82.00          & 74.15          & \textbf{76.93} & \textbf{84.00} & \textbf{0.950} & 0.815          & \textbf{0.911} & \textbf{0.915} & \textbf{0.945} \\ 
\multicolumn{1}{c}{HoHoNet\cite{Sun_2021_HoHoNet}}       & 82.64          & 82.16          & 73.65          & 69.26          & 79.88          & 85.26          & 84.81          & 75.59          & 70.98          & 82.32          & -   & -   & -   & -   & -   \\
\multicolumn{1}{c}{LED$^2$-Net\cite{Wang_2021_LED2Net}}      & 84.22          & \textbf{83.22} & \textbf{76.89} & 70.09          & 81.52          & 86.91          & \textbf{85.53} & \textbf{78.72} & 71.79          & 83.91          & -   & -   & -   & -   & -   \\ 
\multicolumn{1}{c}{\ \ \ \textbf{DMH-Net (Ours)}\ \ \ } & \textbf{84.39} & 80.22          & 66.15          & 64.46          & 78.97          & \textbf{86.94} & 82.31          & 67.99          & 66.2& 81.25  & 0.949          & \textbf{0.951} & 0.838          & 0.864          & 0.925      \\
\bottomrule
\end{tabular}%
}
\caption{Quantitative results of non-cuboid room layout estimation evaluated on the Matterport 3D dataset~\cite{Matterport3D}.}
\label{tab:matter}
\end{table*}

\subsection{Ablation Study}\label{sec:ablation_result} 
In this section, we perform an ablation study to demonstrate the effect of the major designs in our method.

\noindent\textbf{Choice of Manhattan Line Types}
We first study the necessity of the three types of lines in the DMHT.
We conduct experiments with each of the types of line detection omitted, respectively. The results are summarized in Tab~\ref{tab:abla-line}. All of the types of line detection are important as disabling any of each results in an obvious performance loss. Among them, horizontal lines detection is more important, as horizontal lines make up the ceiling and floor wireframe which is essential to the initialization in the post-processing procedure.

\noindent\textbf{Cubemap and DMHT}
We then verify whether the cubemap representation and DMHT helps to recover 3D room layout.
We analyze the effectiveness of our full pipeline~(\ie \uppercase\expandafter{\romannumeral3}) with another two ablation variants in~\cref{tab:rebuttal_ana}.
The variation (\uppercase\expandafter{\romannumeral1}) encodes feature from cubemap as ours but replaces decoder from DMHT to HorizonNet-like LSTM~\cite{sun2019horizonnet}. 
The significant performance drop without DMHT in (\uppercase\expandafter{\romannumeral1}) indicates the necessity of DMHT in our full pipeline.
We also compare the deep Hough feature against classical Hough Transforms line detectors, please refer to supplementary materials for more details.
The variation (\uppercase\expandafter{\romannumeral2}) directly extracts features from the panorama image then apply E2P transform to feed features into DMHT. It shows the effectiveness of cubemap representation since the distortion in equirectangular features can degrade performance. 
These ablations between (\uppercase\expandafter{\romannumeral1}, \uppercase\expandafter{\romannumeral2}) and (\uppercase\expandafter{\romannumeral3}) provide evidence that the performance gain owe to the use of cubemap and DMHT.

\begin{table}[!b]
\centering
\resizebox{.7\textwidth}{!}{%
\begin{tabu} to 0.85 \textwidth {X[2c]X[2c]X[2c]X[3c]X[3c]X[3c]}

\toprule
\multicolumn{3}{c}{Line Detection}                                              & \multirow{2}{*}{3DIoU} & \multirow{2}{*}{CE} & \multirow{2}{*}{PE} \\ \cmidrule(lr){1-3}
$\mathbf{H}$ & $\mathbf{V}$ & $\mathbf{C}$ &        &     &     \\
\midrule
\texttimes          & \checkmark         & \checkmark        &     43.58         &    3.69      &  11.91   \\
\checkmark           &\texttimes         &   \checkmark             &   83.28                     &   0.89      &     2.44    \\
\checkmark           & \checkmark         &     \texttimes           & 84.15                  & 0.79                & \textbf{1.96}                \\ 
\checkmark           & \checkmark         &   \checkmark             & \textbf{85.48}         & \textbf{0.73}       & \textbf{1.96}       \\
\bottomrule
\end{tabu}
}
\caption{Ablation study on the impact of different types of line detection on the PanoContext dataset, in which $\mathbf{H}$ means horizontal line detection, $\mathbf{V}$ means vertical line detection and $\mathbf{C}$ means detection of lines passing the center.}
\label{tab:abla-line}
\end{table}

\begin{table}[!b]
\centering
\resizebox{.7\textwidth}{!}{%
\begin{tabu} to 0.85 \textwidth {X[1.5c]X[2c]X[2c]X[c]X[c]X[c]}
\toprule
\multirow{2}{*}{Method} & \multicolumn{2}{c}{encoder}                     & \multicolumn{2}{c}{decoder}      & \multirow{2}{*}{3D IoU} \\
\cmidrule(lr){2-3}\cmidrule(lr){4-5}
& {cuebmap} & equirectangular & {DMHT} & RNN &  \\
\midrule
\uppercase\expandafter{\romannumeral1} & {\checkmark} &   & {}  & \checkmark &  81.44 \\
\uppercase\expandafter{\romannumeral2} & {}  & \checkmark & {\checkmark} &   &  83.99  \\
\uppercase\expandafter{\romannumeral3} (Ours) & {\checkmark} &   & {\checkmark} &   & \textbf{85.48} \\
\bottomrule
\end{tabu}
}
\caption{Ablation study on encoder feature representation and decoder choice. We analyze the effectiveness of our pipeline~(\ie \uppercase\expandafter{\romannumeral3}) with another two ablation variants in~\cref{tab:rebuttal_ana}. Our pipeline~(\ie \uppercase\expandafter{\romannumeral3}) achieves best quantitative results.}
\label{tab:rebuttal_ana}
\end{table}

\subsection{Model Analysis}
\noindent\textbf{Encoder Network}
We first investigate how our cubemap representation and DMHT are model agnostic. We choose four different encoder backbones: ResNet-18,34,50\cite{He_2016_ResNet} and DRN-38\cite{yu2017dilated}.
We change backbone for both our network and HorizonNet~\cite{sun2019horizonnet} baseline, and then train and evaluate the networks on the PanoContext dataset.
As shown in Tab~\ref{tab:abla-backbone}, our method is not only effective for the DRN-38 architecture we use, but also consistently outperforms the competitors quantitatively. Ours with Res-50 is even slightly higher than DRN, may owing to deeper layers.

\begin{table}[!h]
\centering
\resizebox{.7\textwidth}{!}{%
\begin{tabu} to 0.85 \textwidth {X[3.0c]X[c]X[c]X[c]X[c]X[c]X[c]}
\toprule
Method  & \multicolumn{3}{c}{Ours}  & \multicolumn{3}{c}{HorizonNet~\cite{sun2019horizonnet}} \\
\cmidrule(lr){1-1}\cmidrule(lr){2-4}\cmidrule(lr){5-7}
Backbone   & 3DIoU & CE & PE     & 3DIoU & CE & PE          \\ 
\midrule
ResNet-18 & \textbf{82.96} & \textbf{0.83} & \textbf{2.18} & 79.02 & 1.52 & 2.77  \\
ResNet-34 & \textbf{83.84} & \textbf{0.87} & \textbf{2.13} & 79.97 & 0.91 & 2.49  \\
ResNet-50 & \textbf{\underline{85.90}} & \textbf{\underline{0.64}} & \textbf{\underline{1.75}} & 82.17 & 0.76 & 2.20  \\
DRN-38    & \textbf{85.48} & \textbf{0.73} & \textbf{1.96} & 81.52 & 0.79 & 2.19  \\
\bottomrule
\end{tabu}
}
\caption{Model analysis of the model agnostic generalization capability of DMH-Net on the PanoContext dataset. (Bold and underline indicates better than other competitors and the best among all, respectively.) Our DMH-Net consistently outperforms the competitors quantitatively when changing different backbones.}
\label{tab:abla-backbone}
\end{table}

\begin{figure}[!b]
	\centering
	\includegraphics[width=0.85\columnwidth]{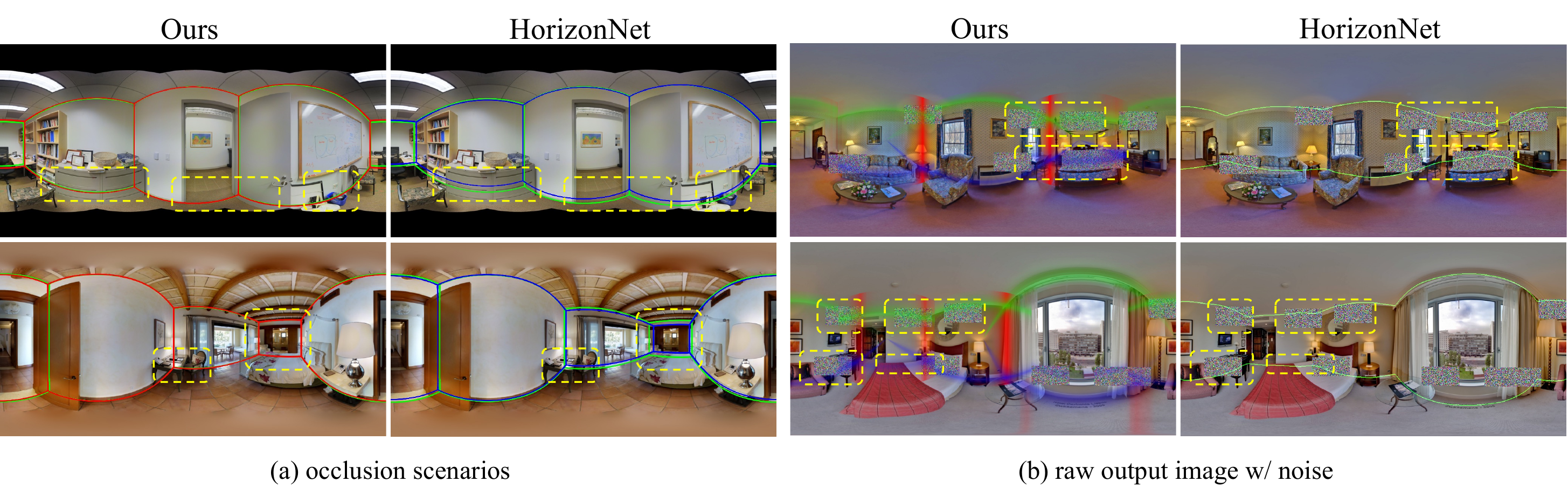}
	\caption{Qualitative comparison in occlusion scenarios. (a) shows the result after post-processing, (b) shows the raw output of the network with challenging noise.}
	\label{fig:fig_occlusion} 
\end{figure}

\noindent\textbf{Robustness to occlusion}
We then test how our method outperforms competitors w.r.t. occlusion scenarios.
We provide qualitative comparison in~\cref{fig:fig_occlusion}(a). 
As illustrated, our method has more accurate prediction results in regions where walls are occluded by other clutter.
We further set up a more challenging experiment. We manually add challenging noise patches, and then test the network without re-training. 
As shown in~\cref{fig:fig_occlusion}(b), the raw output of ours is more reasonable than competitors in noisy regions.
Moreover, quantitative results on the PanoContext dataset also show that our method achieves a 3D IoU of 84.39 while HorizonNet is 78.74.

\begin{figure}[!h]
	\centering
	\includegraphics[width=1\columnwidth]{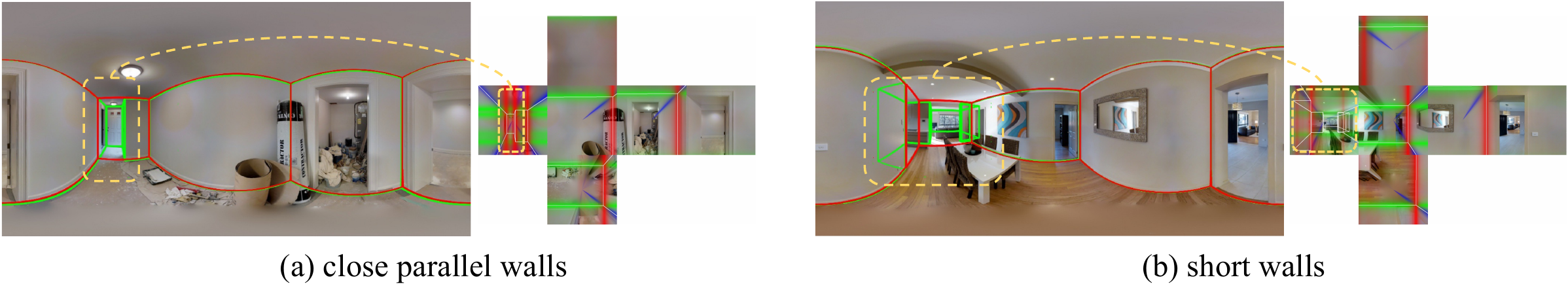}
	\caption{Non-cuboid results explanations.}
	\label{fig:fig_limit} 
\end{figure}

\noindent\textbf{Limitations}
We show two examples for analyzing our performance limitations on non-cuboid data.
\cref{fig:fig_limit}(a) shows that our method may suffer from discriminating two very close walls.
\cref{fig:fig_limit}(b) shows low line confidence due to ``Vote Splitting''. The probability peaks of short lines~(in leftmost tile) are less significant than long ones. The short ones may not have enough feature bins to vote for.
These cases may be improved by exploring adding interactions between tiles and a improved version of DMHT, which will be our future works.

\section{Conclusion}
In this paper, we introduced a new method for estimating 3D room layouts from a cubemap of panorama image by leveraging DMHT. The proposed method detects horizontal lines, vertical lines and lines passing the image center on the cubemap and applies post-processing steps to combine the line prediction results to get the room layout. The learnable Deep Hough transform enables the network to capture the long-range geometric pattern and precisely detect the lines in the wireframe of the room layout. Both quantitative and qualitative results demonstrate that our method achieves better results for cuboid room estimation and comparable results for non-cuboid room estimation with recent state-of-the-art methods in prediction accuracy and performance.
For future work, combining with fully differentiable post-processing or Atlanta world assumption are some of the practical directions to explore.

 \ \\\noindent\textbf{Acknowledgements}
This work was supported by  National Natural Science Funds of China (No. 62088102, 62021002) and ByteDance Research Collaboration Project.
\ULforem

\renewcommand\thesection{\Alph{section}}
\renewcommand\thesubsection{\thesection.\arabic{subsection}}
\newpage
\section*{\centering \Large Supplementary Materials}
\setcounter{section}{0}
\section{Proof of the Manhattan Lines Orientation in Cubemap}
\label{sec:proof}

\noindent{\textbf{Proposition.}} 
Given a panoramic image of a 3D room, if the following two assumptions holds:
\normalem
\begin{enumerate}
    \item The Manhattan world assumption, \ie, all of the walls, the ceiling and the floor must be perpendicular to each other, and all of the intersection lines of them must be parallel with one of the coordinate axes of some orthogonal coordinate space (named Manhattan space).
    \item The input image must be aligned, \ie, the camera of each cubemap tile faces precisely to one of the walls, and its optical axis is parallel with one of the coordinate axes of the Manhattan space.
\end{enumerate}
\ULforem
and the cubemap of the panoramic images is generated by making E2P transform six times with azimuth angle $u=0^{\circ},90^{\circ},180^{\circ},-90^{\circ},0^{\circ},0^{\circ}$ and elevation angle $v=0^{\circ},0^{\circ},0^{\circ},0^{\circ},90^{\circ},-90^{\circ}$ respectively and FoV$=90^{\circ}$ for all cubemap tiles,
then for any of the lines in the wireframe of the 3D room, it must be either a horizontal line ($\theta=0$), a vertical line ($\theta=\pi/2$) or a line passing the center ($\rho=0$) in the cubemap tiles.

\subsubsection{\textbf{(i) Preliminary. Coordinate Definition.}} 
\begin{figure}[!h]
	\centering
	\includegraphics[width=0.7\columnwidth]{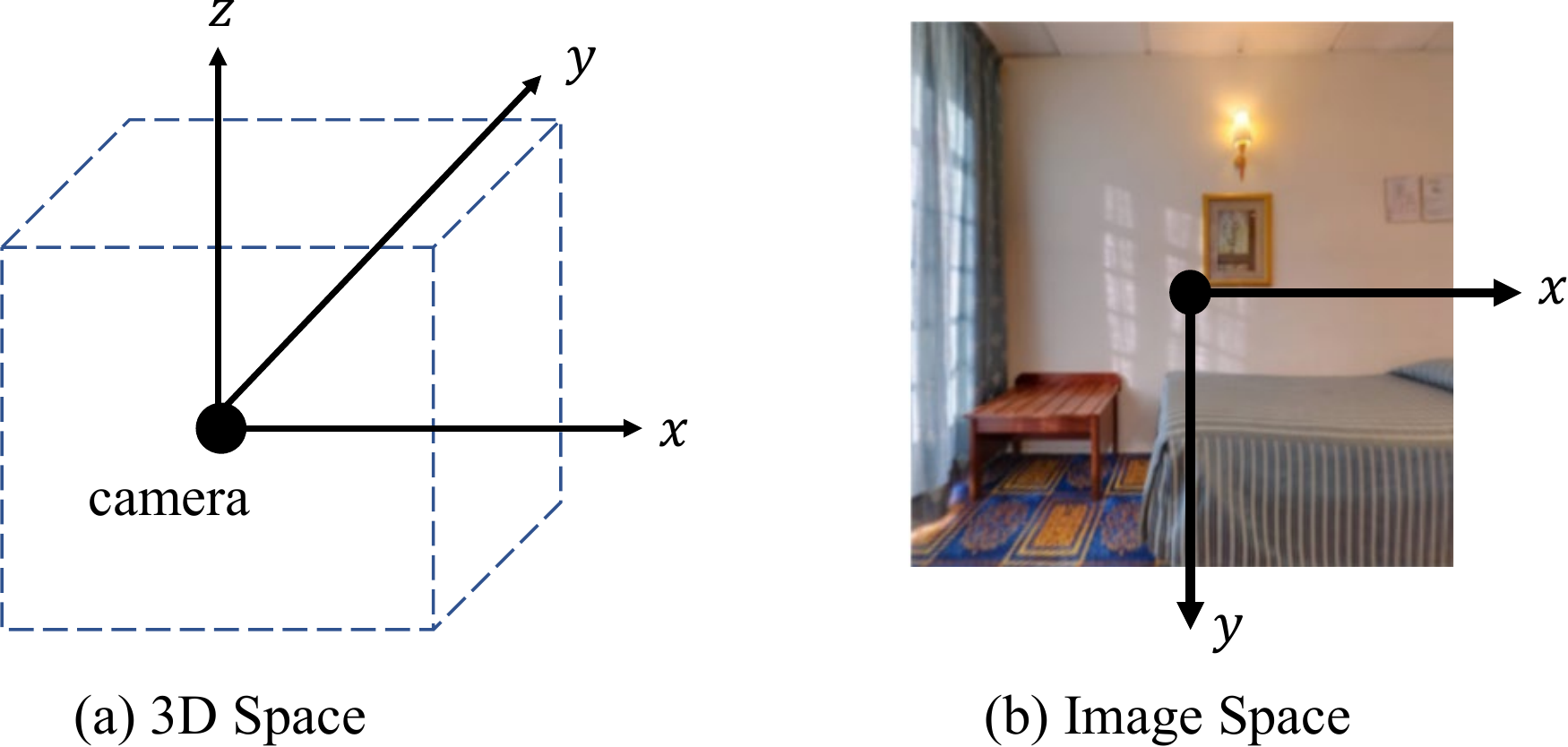}
	\caption{Coordinate space definition of 3D space and image space.}
	\label{fig:coord_def} 
\end{figure}

Two coordinate space is defined: the 3D space and the image space, as \cref{fig:coord_def} shows. In the 3D space, the position of the panoramic camera is the origin, the $y$ axis is along the optical axis of the camera, the $z$ axis points to the ceiling vertically, and the $x$ axis is orthogonal to the $y$ axis points to the right. In the image space, the center of the image is the origin, the $x$ axis points to the right and the $y$ axis points to the bottom.

\subsubsection{\textbf{(ii) Proof. Coordinate Transform from the 3D Space to the Image Space.}}
Given a coordinate $p$ in the 3D space, to get its coordinate $q$ in the image space of a specific cubemap view, we can first apply the observation transform:
\begin{equation}\label{formula:1}
    r = (M_{u}M_{v})^{-1} \cdot p
\end{equation}
in which $M_{u}$ is the rotation matrix representing rotation along the $z$ axis by angle $u$, and $M_{v}$ is the rotation matrix representing rotation along the $x$ axis by angle $v$:
\begin{equation}\label{formula:2}
    \begin{aligned}
    M_{u}&=\begin{bmatrix}
\cos(u) & -\sin(u) & 0 \\ 
\sin(u) & \cos(u)  & 0 \\ 
0 & 0 & 1 
\end{bmatrix}\\
    M_{v}&=\begin{bmatrix}
1 & 0 & 0 \\
0 & \cos(v) & -\sin(v) \\ 
0 & \sin(v) & \cos(v)   
\end{bmatrix}
    \end{aligned}
\end{equation}
Then we can apply the projection:
\begin{equation}\label{formula:3}\begin{aligned}
    q_x &= \frac{r_x}{r_y}\cdot \cot(\alpha_H) \cdot \frac{w}{2} \\
    q_y &= \frac{-r_z}{r_y}\cdot \cot(\alpha_V) \cdot \frac{h}{2}
\end{aligned}\end{equation}
in which $q=(q_x, q_y)^\top$ is the coordinate in the image space, $r_x, r_y$ is the $x, y$ components of $r$, $\alpha_H, \alpha_V$ is the FoV in horizontal and vertical direction respectively, and $h, w$ is the height and width of the image. 

Since $\alpha_H, \alpha_V = 90^{\circ}$, from Eq.~(\ref{formula:1})~(\ref{formula:2})~(\ref{formula:3}), we have
\begin{equation}\hspace{-0.1cm}
\label{formula:4}\begin{aligned}
    \medmath{q_x} &= \medmath{\frac{w}{2} \cdot \frac{p_x\cos(u)+p_y\sin(u)}{-p_x\cos(v)\sin(u)+p_y\cos(v)\cos(u)+p_z\sin(v)}} \\
    \medmath{q_y} &= \medmath{\frac{h}{2} \cdot \frac{-p_x\sin(v)\sin(u)+p_y\sin(v)\cos(u)-p_z\cos(v)}{-p_x\cos(v)\sin(u)+p_y\cos(v)\cos(u)+p_z\sin(v)}}
\end{aligned}
\end{equation}

\subsubsection{\textbf{(iii) Conclusion. Manhattan Line
Orientation. }}
Take the cubemap tile at the front of the camera, which corresponds to $u=0^{\circ}, v=0^{\circ}$, as an example. By substituting $u$ and $v$ into \cref{formula:4}, we get
\begin{equation}\hspace{-0.3cm}
\label{formula:5}\begin{aligned}
    q_x &= \frac{w}{2}\cdot\frac{p_x}{p_y} \\
    q_y &= \frac{h}{2}\cdot\frac{-p_z}{p_y}
\end{aligned}\end{equation}

For a horizontal wireframe line which is along the $x$ axis in the 3D space, take any two points $p_1=(p_{x1}, p_{y1}, p_{z1})^\top$ and $p_2=(p_{x2}, p_{y1}, p_{z1})^\top$ on the line (Notice that the two points has the same $y$ and $z$ coordinates). By \cref{formula:5}, we have $q_1=\left(\frac{w}{2}\cdot\frac{p_{x1}}{p_{y1}}, \frac{h}{2}\cdot\frac{-p_{z1}}{p_{y1}}\right)$, $q_2=\left(\frac{w}{2}\cdot\frac{p_{x2}}{p_{y1}}, \frac{h}{2}\cdot\frac{-p_{z1}}{p_{y1}}\right)$. $q_1$ and $q_2$ have the same $y$ and different $x$ coordinate in the image space, so it is a \coloruline{customdeepgreen}{horizontal line} in the cubemap tile. 

For a vertical wireframe line which is along the $z$ axis in the 3D space, take any two points $p_3=(p_{x3}, p_{y3}, p_{z3})^\top$ and $p_4=(p_{x3}, p_{y3}, p_{z4})^\top$. Similarly by \cref{formula:5}, we have $q_3=\left(\frac{w}{2}\cdot\frac{p_{x3}}{p_{y3}}, \frac{h}{2}\cdot\frac{-p_{z3}}{p_{y3}}\right)$, $q_4=\left(\frac{w}{2}\cdot\frac{p_{x3}}{p_{y3}}, \frac{h}{2}\cdot\frac{-p_{z4}}{p_{y3}}\right)$, which have the same $x$ and different $y$ coordinate in the image space, so it is a \colorulinea{red}{vertical line} in the cubemap tile.

Finally, for a horizontal wireframe line which is along the $y$ axis in the 3D space, take any two points $p_5=(p_{x5}, p_{y5}, p_{z5})^\top$ and $p_6=(p_{x5}, p_{y6}, p_{z5})^\top$. Similarly by \cref{formula:5}, we have $q_5=\left(\frac{w}{2}\cdot\frac{p_{x5}}{p_{y5}}, \frac{h}{2}\cdot\frac{-p_{z5}}{p_{y5}}\right)$, $q_6=\left(\frac{w}{2}\cdot\frac{p_{x5}}{p_{y6}}, \frac{h}{2}\cdot\frac{-p_{z5}}{p_{y6}}\right)$. Naturally, we have $\frac{q_{y5}-0}{q_{x5}-0}=\frac{-h\cdot p_{z5}}{w\cdot p_{x5}}=\frac{q_{y6}-0}{q_{x6}-0}$, so the three points $q_5$, $q_6$ and $(0,0)$ lie on the same line, {\normalem \ie} the line passing $q_5$ and $q_6$ must be a \colorulineb{blue}{line that passes the center} of the cubemap tile. 

Similarly, the proposition can also be proved for the other cubemap tiles by substituting the $u$ and $v$ into \cref{formula:4}, and derive the point coordinates in the image space for each of the three directions of lines in the 3D space.

\normalem
\section{Implementation Details}
\subsection{Hough Voting for Lines Passing the Center}
\label{sec:detail_center}
Given a channel of the features of a cubemap tile $\mathbf{X}\in\mathbb{R}^{h \times w}$, in our implementation, the vector generated by Manhattan Hough Transform $\mathbf{C}\in\mathbb{R}^{2(h+w)}$, with each of the bins in $\mathbf{C}$ representing a line which starts at the center ($x=0, y=0$) and ends at a bin on the border of the image ($x=x_0, y=y_0$), in which either $x_0=\pm\frac{w}{2}, y_0\in\mathbb{Z}$ or $x_0\in\mathbb{Z}, y_0=\pm\frac{h}{2}$. Note that in our experiment, $h=w=512$.

For each line, if it intersects the border of the image on the left or the right, \ie $\Delta y < \Delta x$, the value in the bin of the Hough vector $\mathbf{C}$ is the sum of $\frac{w}{2}$ values corresponding to $\frac{w}{2}$ points whose coordinate $x$ is integer. For each integer $x$, we calculate the $y$ coordinate (which may not be integer). If $y\in\mathbb{Z}$, the value of this point is naturally defined as the pixel $\mathbf{X}[x,y]$. (The brackets ``$[$~$]$'' denote the index of the feature map.) If $y\notin\mathbb{Z}$, the value of this point is defined as the linear interpolation of the two pixel $\mathbf{X}[x,\lfloor y \rfloor]$ and $\mathbf{X}[x,\lceil y \rceil]$. 


When the line intersects the border of the image on the top or the bottom, \ie $\Delta x < \Delta y$, the definition is similar except that the points used for summation is selected by coordinate $y$ being integer.

\subsection{Network Structure}
Here we introduce our network structure in detail. Our Network can be divided into three parts: feature extractor, Manhattan Hough Head and output modules. 

\subsubsection{Feature Extractor.} In the following example, we take DRN-38\cite{yu2017dilated} as the encoder. The structure of the encoder blocks are summarized in \cref{tab:encoder}. 

\begin{table}[t]
\centering
\begin{tabu} to 0.7\columnwidth {X[c]X[2.5,c]X[2,c]}
\toprule
\textbf{Block}  & \textbf{Output Channels} & \textbf{Output Size} \\ \midrule
\texttt{block1} & 32       & $256\times 256$      \\ \midrule
\texttt{block2} & 64       & $128\times 128$         \\ \midrule
\texttt{block3} & 128      & $64\times 64$          \\ \midrule
\texttt{block4} & 256      & $64\times 64$          \\ \midrule
\texttt{block5} & 512      & $64\times 64$          \\ \bottomrule
\end{tabu}
\caption{Summary of the feature maps output by each block of the encoder.}
\label{tab:encoder}
\end{table}

\subsubsection{Manhattan Hough Head.}
We extract multi-scale features from the encoder. We obtain five feature maps of different scales and feed them into the corresponding Manhattan Hough Head module.
The network contains five separate Manhattan Hough Heads.
In \cref{tab:hough_head}, we take the output of block1 of the encoder, \ie \texttt{encoder.block1}, as an example to introduce the structure of the Manhattan Hough Head. The final three layers, \texttt{conv1d\_H.2}, \texttt{conv1d\_V.2} and \texttt{conv1d\_C.2} generates the three output feature vectors $\mathbf{H}$, $\mathbf{V}$ and $\mathbf{C}$ of the Manhanttan Hough Head respectively.

\subsubsection{Output Modules.}
The final module processes the three types of feature vectors of $\mathbf{H}$, $\mathbf{V}$ and $\mathbf{C}$ to obtain probabilities. We take the output module corresponding to $\mathbf{H}$ as an example to introduce the detailed structure.
Since feature vectors from each Manhattan Hough Head have different size, they must be upsampled before being concatenated. Specifically, the channels of the output by each Manhattan Hough Head are $16, 32, 64, 128, 256$ respectively. Denote $\mathbf{H}^{\uparrow}_{n}$ is the $\mathbf{H}$ feature vectors output by the $n$-th Manhattan Hough Head after being upsampled, the structure of the output module is shown in \cref{tab:output}. Note that there are 
another two separate output modules for $\mathbf{V}$ and $\mathbf{C}$
, whose structures are the same except for output size. 

\begin{table*}[!t]
\centering

\begin{tabu} to \textwidth {X[4,c]X[5,c]X[3,c]X[3,c]X[2,c]X[1,c]X[1,c]X[2,c]}
\toprule
\textbf{Layer}             & \textbf{Input}             & \textbf{Channels}          & \textbf{Output Size}     & $\mathbf{k}$ & $\mathbf{p}$ & \textbf{BN}           & \textbf{Act.} \\ \midrule
\texttt{conv2d\_H}         & \texttt{encoder.block1}    & 32$\rightarrow$16 & $256\times 256$ & $1\times1$   & 0            & $\checkmark$ & ReLU \\ \midrule
\texttt{conv2d\_V}         & \texttt{encoder.block1}    & 32$\rightarrow$16 & $256\times 256$ & $1\times1$   & 0            & $\checkmark$ & ReLU \\ \midrule
\texttt{conv2d\_C}         & \texttt{encoder.block1}    & 32$\rightarrow$16 & $256\times 256$ & $1\times1$   & 0            & $\checkmark$ & ReLU \\ \midrule
$\mathcal{MH}$\_H & \texttt{conv2d\_H}         & 16$\rightarrow$16                & 256             &              &              &              &      \\ \midrule
$\mathcal{MH}$\_V & \texttt{conv2d\_V}         & 16$\rightarrow$16                & 256             &              &              &              &      \\ \midrule
$\mathcal{MH}$\_C & \texttt{conv2d\_C}         & 16$\rightarrow$16                & 1024            &              &              &              &      \\ \midrule
\texttt{conv1d\_H.1}       & $\mathcal{MH}$\_H & 16$\rightarrow$16                & 256             & $3$          & 1            & $\checkmark$ & ReLU \\ \midrule
\texttt{conv1d\_V.1}       & $\mathcal{MH}$\_V & 16$\rightarrow$16                & 256             & $3$          & 1            & $\checkmark$ & ReLU \\ \midrule
\texttt{conv1d\_C.1}       & $\mathcal{MH}$\_C & 16$\rightarrow$16                & 1024            & $3$          & 1            & $\checkmark$ & ReLU \\ \midrule
\texttt{conv1d\_H.2}       & \texttt{conv1d\_H.1}       & 16$\rightarrow$16                & 256             & $3$          & 1            & $\checkmark$ & ReLU \\ \midrule
\texttt{conv1d\_V.2}       & \texttt{conv1d\_V.1}       & 16$\rightarrow$16                & 256             & $3$          & 1            & $\checkmark$ & ReLU \\ \midrule
\texttt{conv1d\_C.2}       & \texttt{conv1d\_C.1}       & 16$\rightarrow$16                & 1024            & $3$          & 1            & $\checkmark$ & ReLU \\ \bottomrule
\end{tabu}
\caption{\textbf{Manhattan Hough Head Architecture.} For all layers we show the input and the number of channels. For convolution layers, we additionally show the kernel size ($\mathbf{k}$), the padding ($\mathbf{p}$), batch normalization (\textbf{BN}), and the activation function (\textbf{Act.}). The stride of convolution is set to 1.}
\label{tab:hough_head}
\end{table*}

\begin{table*}[!t]
\centering
\begin{tabu} to \textwidth {X[3,c]X[5,c]X[7,c]X[3,c]X[1,c]X[1,c]X[1,c]X[2,c]}
\toprule
\textbf{Layer} & \textbf{Input}   & \textbf{Channels}      & \textbf{Output Size} & \textbf{$\mathbf{k}$} & \textbf{$\mathbf{p}$} & \textbf{BN}  & \textbf{Act.} \\ \midrule
\texttt{concat\_H}      & \fontsize{7pt}{7pt}{$\mathbf{H}^{\uparrow}_{1}, \mathbf{H}^{\uparrow}_{2}, \mathbf{H}^{\uparrow}_{3}, \mathbf{H}^{\uparrow}_{4}, \mathbf{H}^{\uparrow}_{5}$} & \fontsize{1pt}{1pt}{16+32+64+128+256=496} & 512                  &                       &                       &              &               \\ \midrule
\texttt{conv1d\_H.1}    & \texttt{concat\_H}                                                                                                                               & 496$\rightarrow$248    & 512                  & 3                     & 1                     & $\checkmark$ & ReLU          \\ \midrule
\texttt{conv1d\_H.2}    & \texttt{conv1d\_H.1}                                                                                                                             & 248$\rightarrow$248    & 512                  & 1                     & 0                     & $\checkmark$ & ReLU          \\ \midrule
\texttt{conv1d\_H.3}    & \texttt{conv1d\_H.2}                                                                                                                             & 248$\rightarrow$1      & 512                  & 1                     & 0                     &              & Sigmoid       \\ 
\bottomrule
\end{tabu}
\caption{\textbf{Output Modules Architecture.} For all layers we show the input and the number of channels. For convolution layers, we additionally show the kernel size ($\mathbf{k}$), the padding ($\mathbf{p}$), batch normalization (\textbf{BN}), and the activation function (\textbf{Act.}). The stride of convolution is set to 1.}
\label{tab:output}
\end{table*}

\subsection{Post-processing}
\begin{figure*}[!h]
	\centering
	\includegraphics[width=\columnwidth]{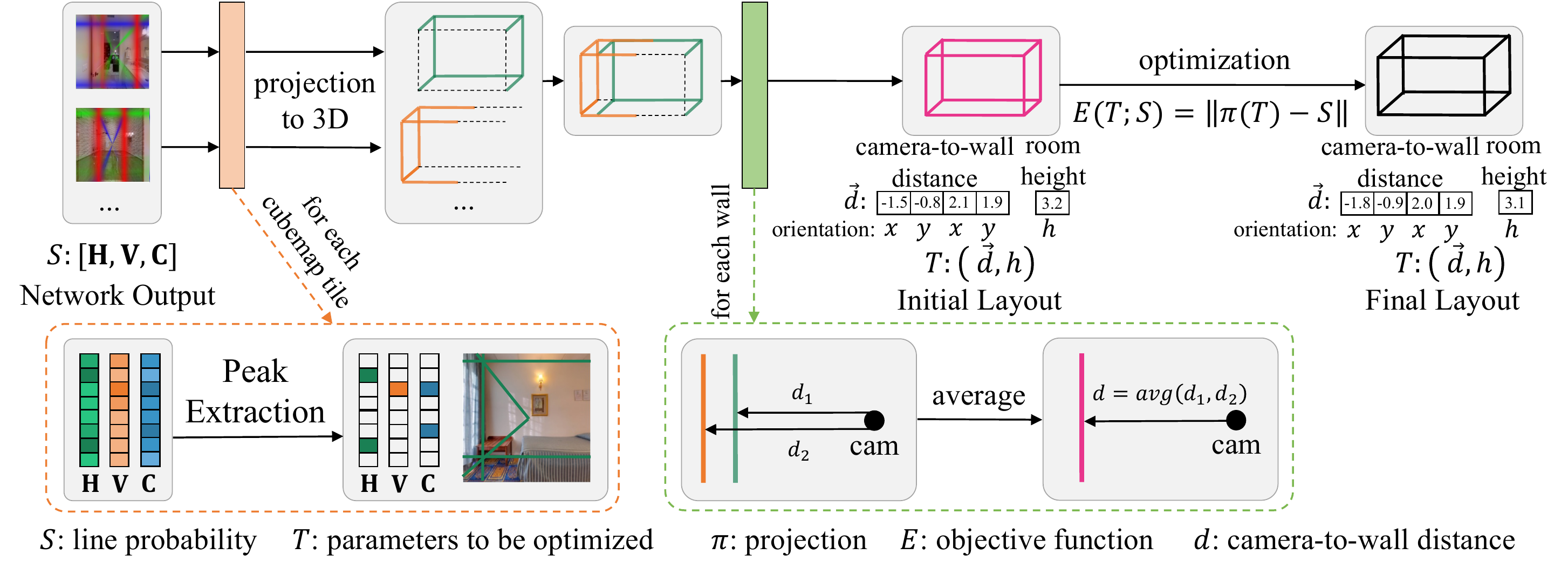}
	\caption{Post-processing pipeline.}
	\label{fig:rebuttal_fig} 
\end{figure*}

We explain the pipeline in Fig.~\ref{fig:rebuttal_fig} that can handle cuboid and non-cuboid rooms. 
\begin{enumerate}
    \item We convert probabilities to lines by finding the prominent peaks using \\{\small{\texttt{scipy.signal.find\_peaks}}}.
    \item Each camera-to-wall distance and room height,~\ie $T$, are optimized here. The parameterized layout $T$, is transformed by projection $\pi$ onto each tile to maximize the overall probability according to the network's line probability output $S$, \ie, $E(T; S)=\|\pi(T)-S\|$.
    \item Necessity: Each wall is overdetermined by the raw output (\ie three types of lines). We convert lines to room height and the camera-to-wall distances $T$, to make the problem determined.
    \item For a MW room, we consider the room walls oriented in an x-y alternation. As the input is axis-aligned, the orientations of other walls can be determined successively, so there is no need for the additional plane normal parameter.
    \item In Tab.~\ref{tab:rebuttal_opt_step}, we add an ablation of different optimization steps to demonstrate the necessity of post-processing.
\end{enumerate}

\begin{table}[!h]
\centering
\begin{tabu} to 0.7\linewidth {X[2,c]X[2.5,c]X[2.5,c]}
\toprule
\multirow{2}{*}{\#Opt. Steps}  & \multicolumn{2}{c}{3D IOU} \\
\cmidrule(lr){2-3} 
& PanoContext  & Stanford 2D-3D          \\
\midrule
0                         & {83.65}          & 81.66          \\
10                        & {84.72}          & 83.04          \\
50                        & {85.30}          & 84.36          \\
100                & {\textbf{85.48}} & \textbf{84.93} \\
\bottomrule
\end{tabu}
\caption{Ablation on number of post-processing optimize steps.}%
\label{tab:rebuttal_opt_step}
\end{table}

\section{More Baseline}
\subsection{Naive Baseline}
An intuitive idea is to apply the classical Hough transform to line detection and then estimate the room layout. However, our experiments demonstrate that this trivial solution cannot solve the room layout estimation problem effectively.

We propose two baselines based on classical Hough transform: Standard Hough Transform ($\mathcal{HT}$-S)~\cite{hough1962method} and Probabilistic Hough Line Transform ($\mathcal{HT}$-P)~\cite{matas2000robust}. 
$\mathcal{HT}$-P improves $\mathcal{HT}$-S, its output format is line segment instead of the whole line.
Specifically, after getting the cubemap of the panoramic image with E2P transform, we first turn the cubemap into a grayscale image and use Canny~\cite{canny1986computational} to detect edge. Then, the two classical Hough transform methods are performed on the egde detection result to detect lines, with voting threshold 100 for $\mathcal{HT}$-S and 50 for $\mathcal{HT}$-P. Though classical Hough transform detect lines of every direction and offset, we only keep those which are horizontal ($|\tan(\theta)|<0.05$), vertical ($|\cot(\theta)|<0.05$) or passing the center ($\rho<5$). 
Since the lines may be too many to perform post-processing, we filter the lines by categorizing them into 8 groups according to their possible position in the 3D space, and keeping only the line with the highest Hough voting value for each group.

\begin{figure*}[!h]
	\centering
	\includegraphics[width=\textwidth]{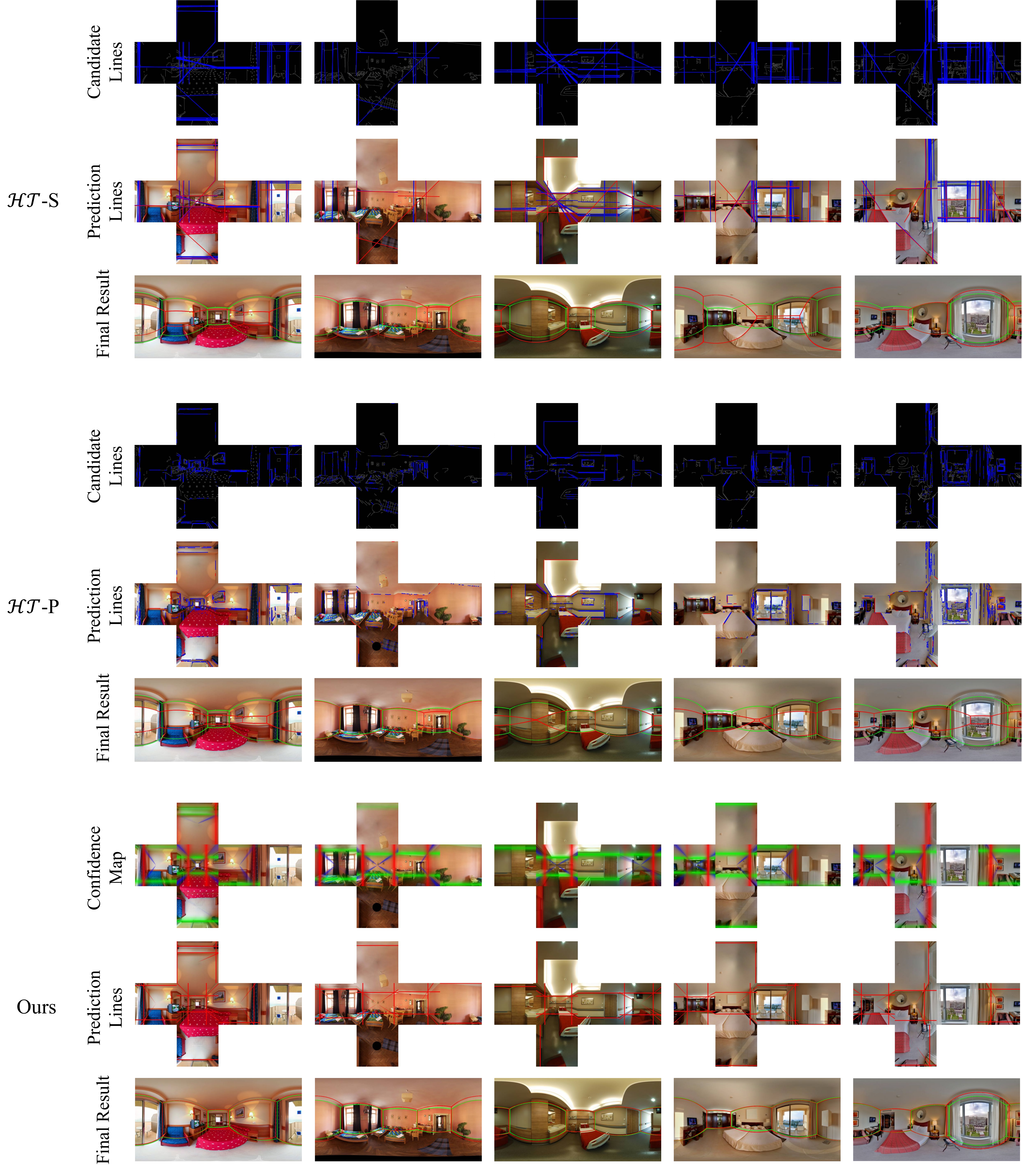}
	\vspace{-1ex}
	\caption{Qualitative results. For $\mathcal{HT}$-S and $\mathcal{HT}$-P: ``Candidate Lines'' visualizes the raw output, and the blue lines are the outputs of Hough transform on the Canny edge map. ``Prediction Lines'' visualizes the lines filtered by the threshold. The blue lines represent all the lines, and the red lines represent the line that meets the threshold. For our method: ``Confidence Map'' visualizes the raw output of network. The colored heatmap represents the probability of the Manhattan lines (described in \cref{sec:proof}). ``Final Result'' visualizes the room layout.  The green lines are ground truth layout while the red lines are estimated.}
	\label{fig:naive} 
\end{figure*}

The quantitative result in the PanoContext dataset\cite{zhang2014panocontext} is shown in \cref{tab:naive}. The qualitative result is shown in \cref{fig:naive}. 
Though classical Hough transform methods are possible to detect the intersection lines of the walls in the room, they may detect more false-positive lines. The lines not only have no benefits to layout estimation but also make the post-processing module fail to give correct estimation result. In contrast, our proposed DMH-Net uses Deep Manhattan Hough Transform which can learn semantic information from the image and make more accurate detection of the intersection lines of the walls in the room rather than detecting all lines equally. 

\begin{table}[!h]
\centering
\begin{tabu} to 0.7\linewidth {X[2,c]X[2,c]X[2,c]X[2,c]}
        \toprule
        \multirow{2}{*}{Method}  & \multicolumn{3}{c}{PanoContext} \\
        \cmidrule(lr){2-4} 
            & 3DIoU  & CE & PE          \\
            \midrule
            $\mathcal{HT}$-S~\cite{hough1962method}              & {23.83}          & {7.01}          & 21.49         \\
            
            $\mathcal{HT}$-P~\cite{matas2000robust}                & {21.78}          & {6.50}          & 17.20          \\
            
            \textbf{DMH-Net} & {\textbf{85.48}} & {\textbf{0.73}} & \textbf{1.96} \\ 
        \bottomrule
    \end{tabu}

\caption{Quantitative results of cuboid room layout estimation evaluated on the PanoContext dataset~\cite{zhang2014panocontext}. CE means Corner Error, and PE means Pixel Error.}
\label{tab:naive}
\end{table}

\subsection{Baseline Discussion}
In the main text, we compare multiple baseline algorithms. To ensure fair comparisons, we use the train/test split that most algorithms use commonly. Since the algorithms proposed in the Zou~\etal~\cite{zou2021_layoutv2} and LGT-Net~\cite{jiang2022lgt} both use additional data in PanoContext and Stanford 2D-3D experiments, we do not quantitatively compare such methods for fairness.

\section{More Model Analysis}
\subsection{Voting bin size}
The number of bins, \ie bin size, affects the performance. The line predictions will be coarser when the number of bins is reduced.
We add an ablation study with 1/2 and 1/4 bins in \cref{tab:rebutal_bin_size}.

\begin{table}[]
    \centering
    \begin{tabu} to 0.7\linewidth {X[2,c]X[2.5,c]X[2.5,c]}
        \toprule
        \multirow{2}{*}{\#Bins}  & \multicolumn{2}{c}{3D IOU} \\
\cmidrule(lr){2-3} 
& PanoContext  & Stanford 2D-3D          \\
\midrule
        $\times$1                  & {\textbf{85.48}}       & \textbf{84.93}          \\
        $\times$1/2                      & {84.73}       & 83.43          \\
        $\times$1/4                      & {83.77}       & 78.81          \\
        \bottomrule
    \end{tabu}
    \caption{Ablation on number of bins.}%
  \label{tab:rebutal_bin_size}
\end{table}

\subsection{Speed analysis}
We provide time profiling per cuboid room sample on a PC with an Intel i7-8700 CPU and a single NVIDIA 2080Ti GPU. Pre-processing: 0.297s. Network inference: 0.155s. Post-processing: 0.643s.

\section{Failure Cases and Limitations}
Two failure cases are given in \cref{fig:bad_case}. Though our method can effectively estimate the 3D room layout for most cases, it also has some limitations. Firstly, our model predict confidence vectors which varies continuously in the Hough space.
Thus, it is challenge for our method to distinguish two very near parallel lines in the cubemap tile, as is shown in the cumbmap tile at line 2, column 4 of \cref{fig:bad_case}~(a). Secondly, our method is still lack of interaction between different cubemap tiles during training. Finally, when the wall-wall intersection line is less salient than other lines on the wall, our method may predict incorrect lines, which leads to a layout estimation error. For example, in \cref{fig:bad_case}~(b), our method predict the lines between the blue part and the white part as the wall-ceiling intersection lines of the room.

\begin{figure}[!h]
\centering
\includegraphics[width=0.6\textwidth]{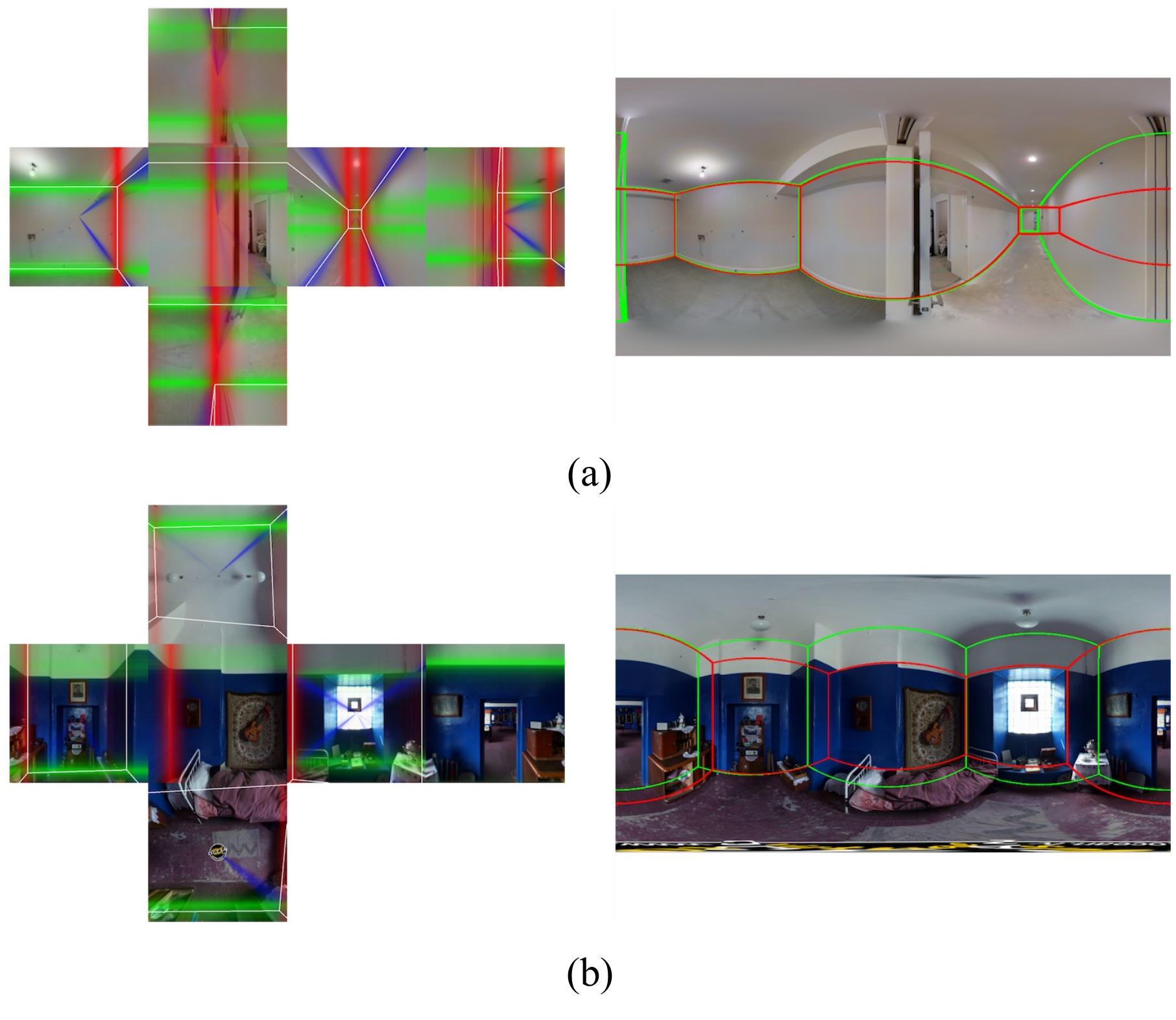}
\label{fig:bad_case}
\caption{Failure cases of our model. (a) is a non-cuboid sample in the Matterport3D dataset~\cite{Matterport3D}, (b) is a cuboid sample in the PanoContext dataset~\cite{zhang2014panocontext}. The green, red and blue lines heatmap in cubemap represents the probabilities of Manhattan line as described in \cref{sec:proof}. 
The thin white line in cubemap represents ground truth.
The green lines in equirectangular panorama are ground truth layout and red lines are predictions.}
\label{fig:bad_case}
\end{figure}


\section{More Qualitative Results}
\subsection{Qualitative Results Compared with Baselines}
\begin{figure*}[!h]
	\centering
	\vspace{-0.5cm}
	\includegraphics[width=\textwidth]{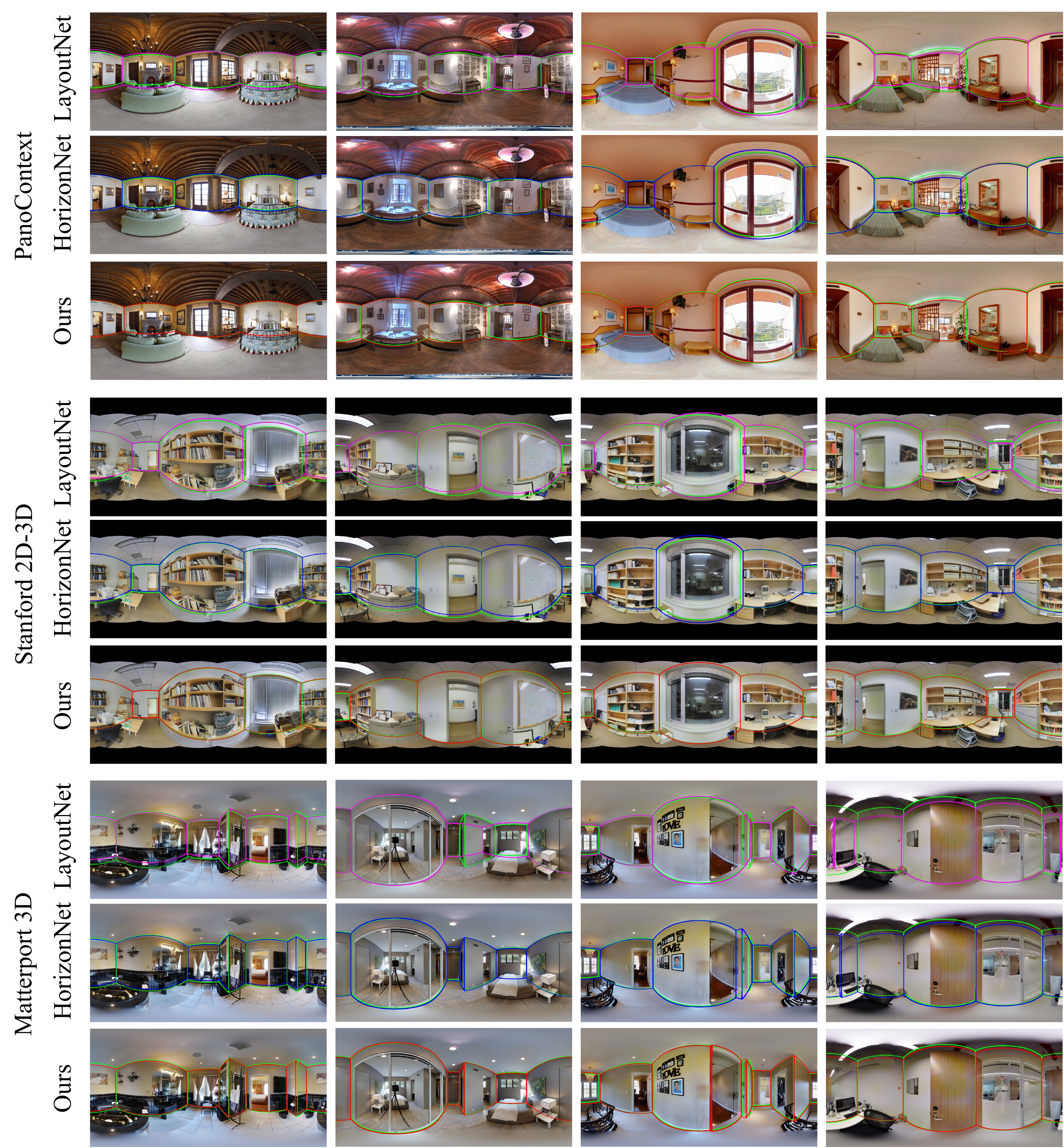}
	\vspace{-1ex}
	\caption{More qualitative results of both cuboid and non-cuboid layout estimation~(1). The green lines are ground truth layout while the pink, blue and red lines are estimated by LayoutNet~\cite{zou2018layoutnet}, HorizonNet~\cite{sun2019horizonnet} and our DMH-Net.}
	\label{fig:hor_lay_cmp1} 
\end{figure*}

\begin{figure*}[!h]
	\centering
	\vspace{-0.5cm}
	\includegraphics[width=\textwidth]{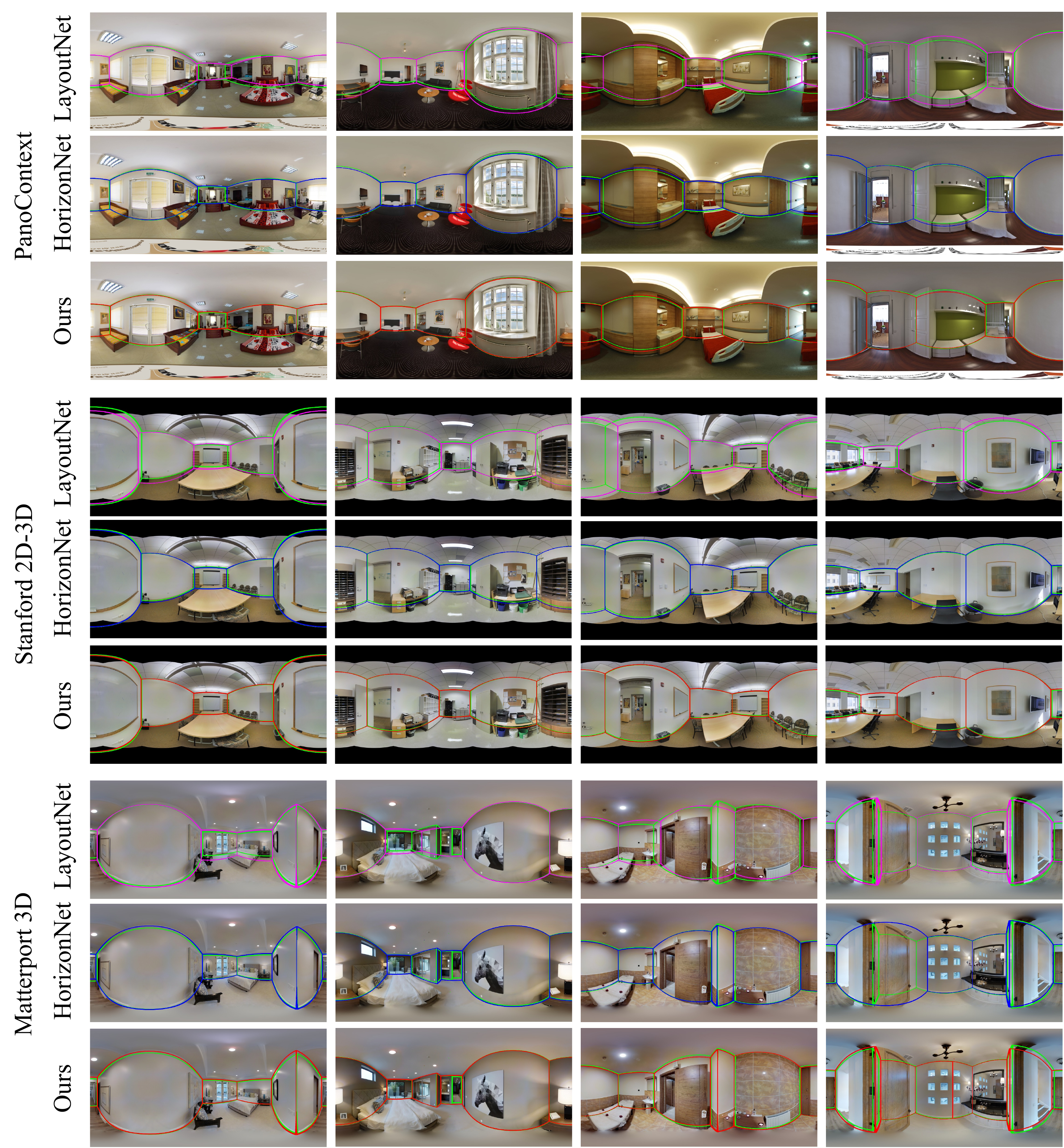}
	\vspace{-1ex}
	\caption{More qualitative results of both cuboid and non-cuboid layout estimation~(2). The green lines are ground truth layout while the pink, blue and red lines are estimated by LayoutNet~\cite{zou2018layoutnet}, HorizonNet~\cite{sun2019horizonnet} and our DMH-Net.}
	\label{fig:hor_lay_cmp2} 
\end{figure*}

For thorough comparison with our baselines, we show additional qualitative results.
As illustrated in Fig.~\ref{fig:hor_lay_cmp1} and Fig.~\ref{fig:hor_lay_cmp2}, we provide the qualitative comparison results with LayoutNet~\cite{zou2018layoutnet,zou2021_layoutv2} and HorizonNet~\cite{sun2019horizonnet} on Stanford 2D-3D dataset~\cite{armeni2017joint}, PanoContext dataset~\cite{zhang2014panocontext} and Matterport3D dataset~\cite{Matterport3D}.

\subsection{Room Layout Comparison with Other Baselines}
\begin{figure*}[!h]
	\centering
	\vspace{-0.5cm}
	\includegraphics[width=\textwidth]{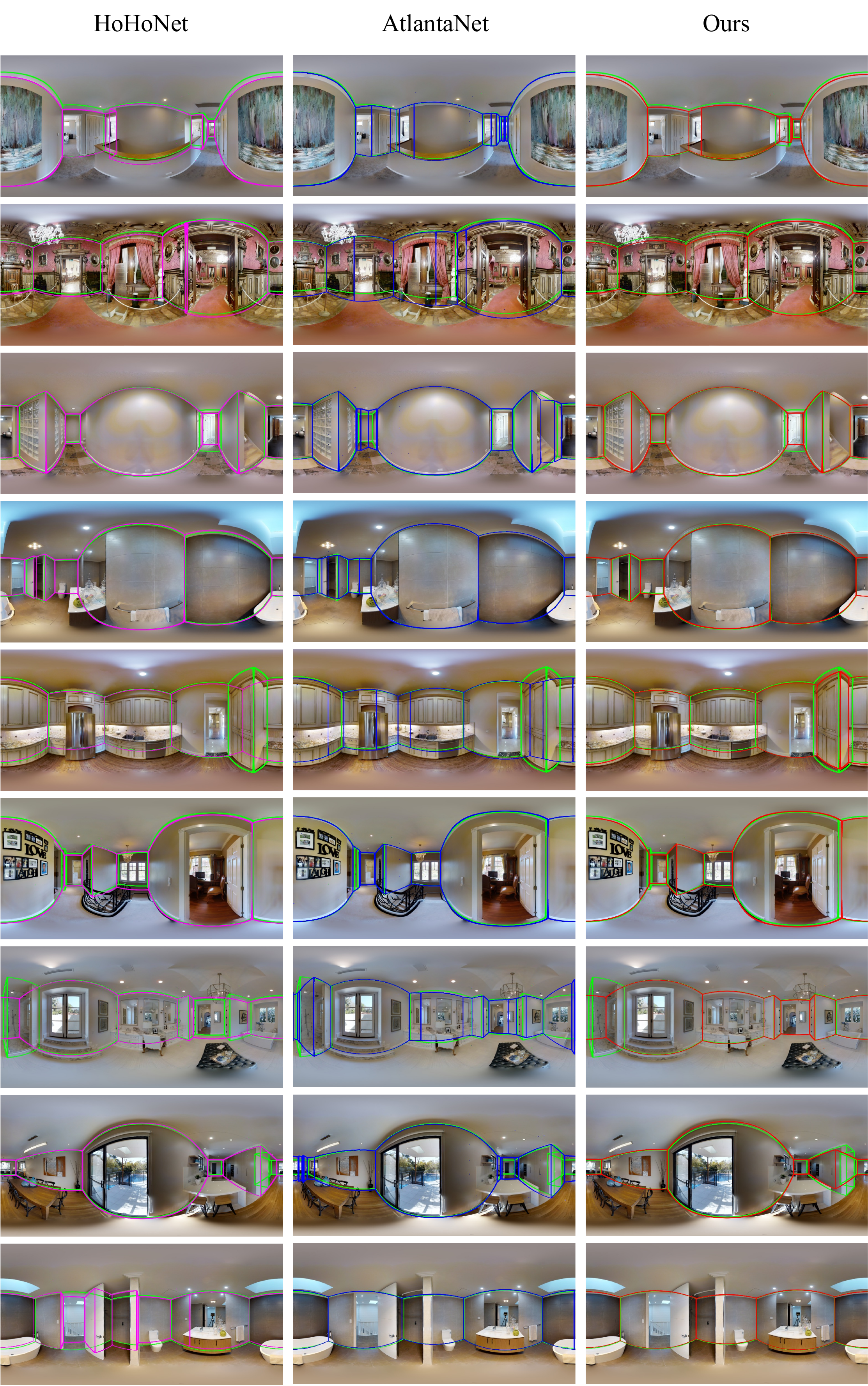}
	\vspace{-1ex}
	\caption{The qualitative results of Matterport3D~\cite{Matterport3D}. The green lines are ground truth layout while the pink, blue and red lines are estimated by HoHoNet~\cite{Sun_2021_HoHoNet}, AtlantaNet~\cite{pintore2020atlantanet} and our DMH-Net.}
	\label{fig:hoho_atl_cmp} 
\end{figure*}

In addition, we provide the qualitative comparison results with HoHoNet~\cite{Sun_2021_HoHoNet} and AtlantaNet~\cite{pintore2020atlantanet} on the Matterport3D dataset~\cite{Matterport3D}, as illustrated in Fig.~\ref{fig:hoho_atl_cmp}.

\subsection{3D Visualization}
\begin{figure*}[!h]
	\centering
	\includegraphics[width=\textwidth]{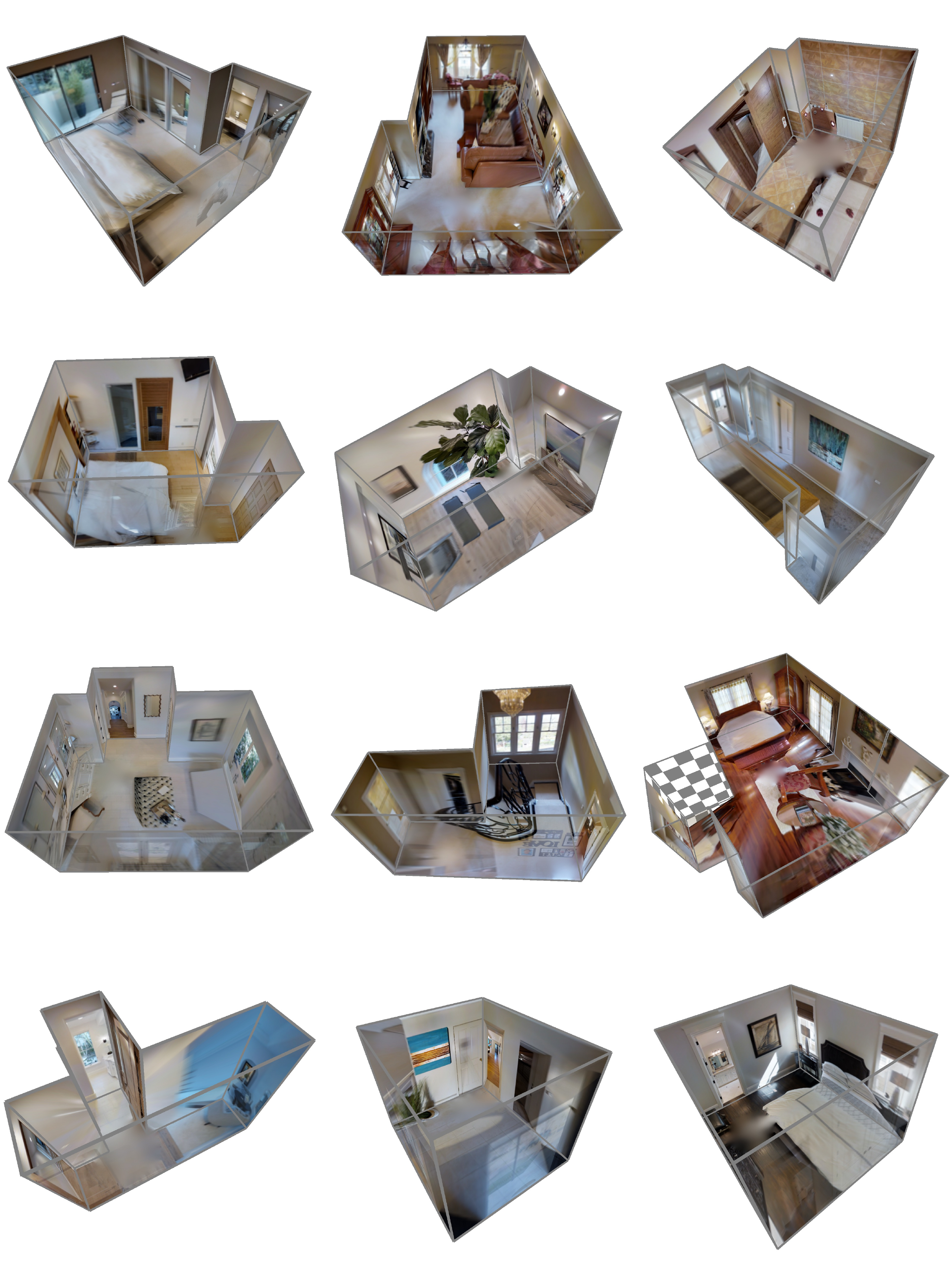}
	\vspace{-1ex}
	\caption{The 3D visualization results of Matterport3D dataset~\cite{Matterport3D}.}
	\label{fig:3d1} 
\end{figure*}

\begin{figure*}[!h]
	\centering
	\includegraphics[width=\textwidth]{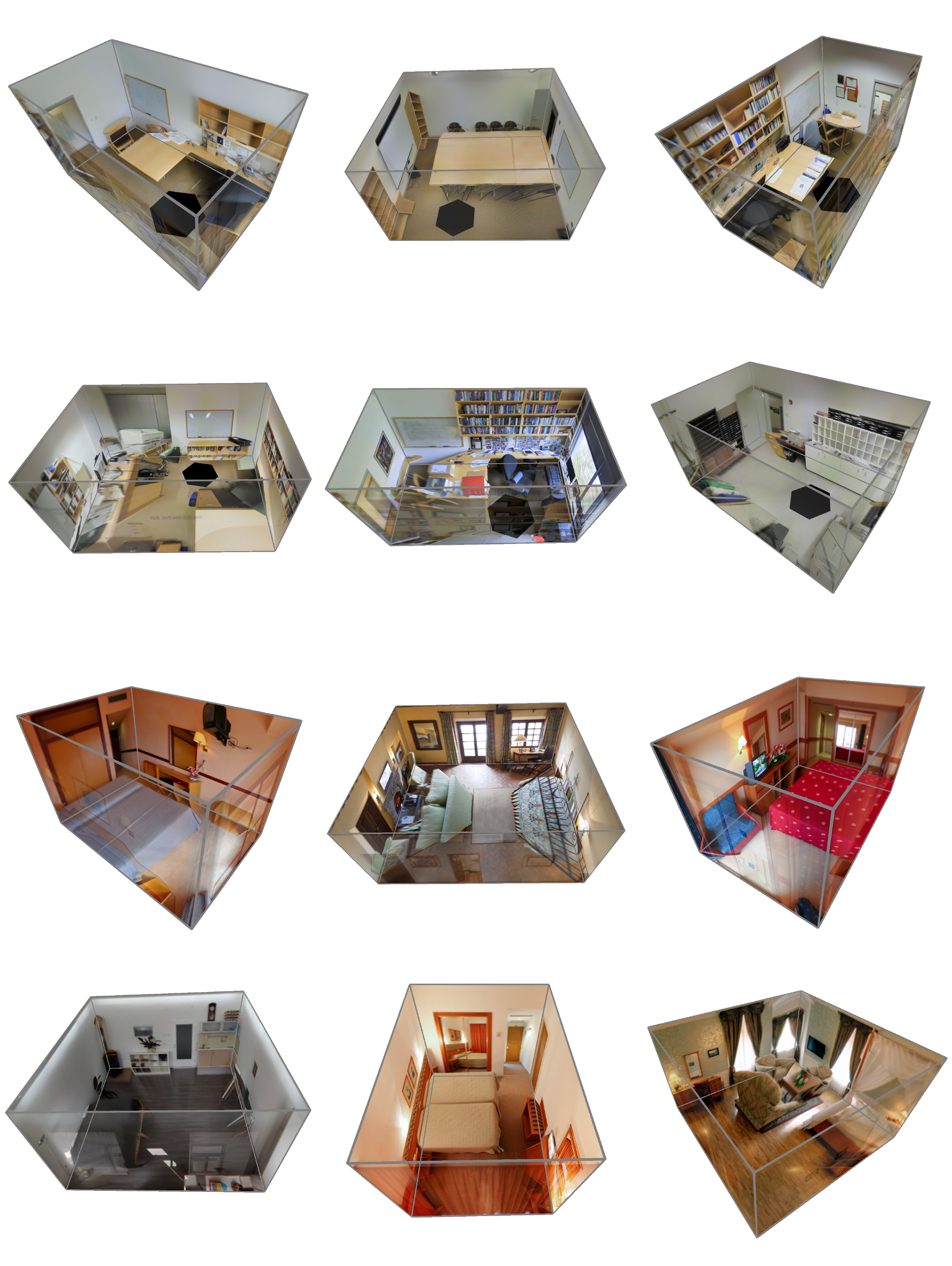}
	\caption{The 3D visualization results of PanoContext dataset~\cite{zhang2014panocontext} and Stanford 2D-3D dataset~\cite{armeni2017joint}.}
	\label{fig:3d2} 
\end{figure*}

We also show 3D layout visualizations of our predictions to demonstrate the performance of our proposed method. Specifically, Fig.~\ref{fig:3d1} shows the results of Matterport3D dataset~\cite{Matterport3D}, and Fig.~\ref{fig:3d2} shows the results of Stanford 2D-3D dataset~\cite{armeni2017joint} and PanoContext dataset~\cite{zhang2014panocontext}.

\clearpage
%
%
\bibliographystyle{splncs04}
\bibliography{egbib}
\end{document}